\documentclass[runningheads]{llncs}

\usepackage{eccv}

\usepackage{eccvabbrv}

\usepackage{graphicx}
\usepackage{booktabs}
\usepackage{graphicx}
\usepackage{amsmath}
\usepackage{array}
\usepackage{amssymb}
\usepackage{booktabs}
\usepackage{multirow}
\usepackage{algorithm}
\usepackage{algpseudocode}
\usepackage{amsmath}
\usepackage{bbm}
\usepackage{verbatim}
\usepackage{xcolor}
\usepackage{colortbl}
\usepackage{pifont}
\usepackage{nicematrix}
\usepackage{epstopdf}
\definecolor{mygray}{gray}{0.9}

\usepackage[accsupp]{axessibility}  

\usepackage{hyperref}

\usepackage{orcidlink}

\begin{document}

\title{Multi-RoI Human Mesh Recovery with Camera Consistency and Contrastive Losses} 

\titlerunning{Multi-RoI HMR}

\author{First Author\inst{1}\orcidlink{0000-1111-2222-3333} \and
Second Author\inst{2,3}\orcidlink{1111-2222-3333-4444} \and
Third Author\inst{3}\orcidlink{2222--3333-4444-5555}}

\author{Yongwei Nie\inst{1}\orcidlink{0000-0002-8922-3205} \and
Changzhen Liu\inst{1}\orcidlink{0009-0003-8802-0996} \and
Chengjiang Long\inst{2}\orcidlink{0000-0003-1584-7290} \and
Qing Zhang\inst{3}\orcidlink{0000-0001-5312-2800} \and
Guiqing Li\inst{1}\orcidlink{0000-0002-4598-1522} \and
Hongmin Cai\inst{1}\inst{\thanks{Hongmin Cai is the corresponding author. Email: hmcai@scut.edu.cn}}\orcidlink{0000-0002-2747-7234}}
 
	\authorrunning{Y. Nie et al.}
	
	\institute{South China University of Technology, China \and
		Meta Reality Labs, USA \and
		Sun Yat-sen University, China}

\maketitle

\begin{abstract}
Besides a 3D mesh, Human Mesh Recovery (HMR) methods usually need to estimate a camera for computing 2D reprojection loss. Previous approaches may encounter the following problem: both the mesh and camera are \textit{not} correct but the combination of them can yield a low reprojection loss. To alleviate this problem, we define multiple RoIs (region of interest) containing the same human and propose a multiple-RoI-based HMR method. Our key idea is that with multiple RoIs as input, we can estimate multiple local cameras and have the opportunity to design and apply additional constraints between cameras to improve the accuracy of the cameras and, in turn, the accuracy of the corresponding 3D mesh. To implement this idea, we propose a RoI-aware feature fusion network by which we estimate a 3D mesh shared by all RoIs as well as local cameras corresponding to the RoIs. We observe that local cameras can be converted to the camera of the full image through which we construct a local camera consistency loss as the additional constraint imposed on local cameras. Another benefit of introducing multiple RoIs is that we can encapsulate our network into a contrastive learning framework and apply a contrastive loss to regularize the training of our network. Experiments demonstrate the effectiveness of our multi-RoI HMR method and superiority to recent prior arts. Our code is available at \url{https://github.com/CptDiaos/Multi-RoI}.

\end{abstract}

\section{Introduction}
\label{sec:intro}

Since the seminar work of HMR (Human Mesh Recovery) by \cite{kanazawa2018end}, more and more work attempts to estimate 3D mesh of a human from a single image, for its potential value in VR/AR, virtual try-on and simulative-coaching, etc. 

\begin{figure}[!t]
\centering
 \includegraphics[width=\columnwidth]{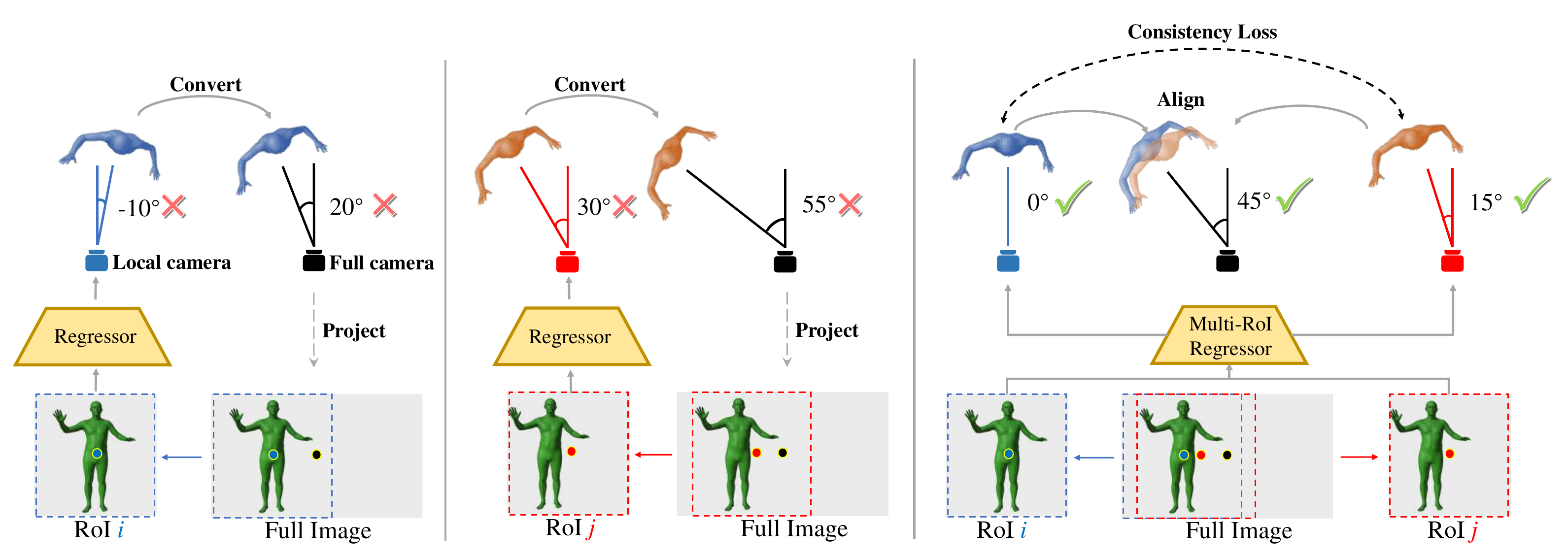}
	\caption{\textbf{(a)} Extracted RoI $i$ is fed to a regressor but it wrongly estimates a local camera which sees the mesh in -10$^\circ$ while the accurate local camera shall see it in 0$^\circ$. Consequently, when further converted to full camera, it will wrongly see the mesh in 20$^\circ$ instead of groundtruth 45$^\circ$. \textbf{(b)} As with RoI $j$, the full camera derived from incorrectly estimated local camera (30$^\circ$) sees the mesh in 55$^\circ$. Both (a) and (b) will mislead the 2D-projection loss to output incorrect 3D mesh due to the false projection. \textbf{(c)} We feed multiple RoIs into the network simultaneously and estimate local cameras of the RoIs. Both local cameras can be converted to the full camera from the perspective of which the 3D mesh should be aligned. We use this observation to establish pairwise consistency losses between local cameras to obtain accurate local cameras (0$^\circ$ and 15$^\circ$).} 
\label{fig:figone}
\end{figure}

Most of previous work, inspired by \cite{kanazawa2018end}, treats this task as a regression problem \cite{zhang2023pymaf,cho2022cross,li2023niki,li2022cliff,kanazawa2018end,wang2023zolly}. They first detect the human from an original full image and use the detected boundingbox to crop the RoI (region of interest) of the human and feed it to a neural network for estimating the target SMPL~\cite{loper2015smpl} mesh together with a local camera. The camera is used to project the mesh to the 2D RoI plane, such that the projected mesh can be compared with 2D evidences (e.g., poses and human joints) in the given RoI to compute the so-called reprojection loss. However, the reprojection loss may be deceived by the mesh and camera. That is, when the mesh and camera are incorrect, their combination may still yield a low reprojection error.

Through the above analysis, we find that without accurate camera for projection during training, 2D-reprojection loss will be misled, making the network learn incorrect mesh configurations (\eg, wrong global orientation or incorrect joint rotations). 
To improve the accuracy of mesh, network needs to estimate more accurate camera parameters. Previous approaches for improving cameras can be classified into two categories. The first kind of methods improve the camera projection model. For example in \cite{wang2023zolly}, the usually adopted weak-perspective camera is replaced with a perspective-distorted camera model with which the distortion in close-up images can be modeled. In \cite{kissos2020beyond,li2022cliff}, the 3D mesh is projected onto the full image and the reprojection loss is computed in the whole view of the full image. This is because the original RoI-view camera has ambiguity on reasoning about the global orientation of the 3D mesh, while the full-view camera model can resolve this ambiguity. Although this kind of methods can reduce the structural error brought by the inappropriate camera model, they cannot guarantee the camera parameters they estimated are accurate, and still cannot prevent the mesh and camera from deceiving the network. The second kind of methods directly design and impose additional constraints on the camera parameters. For example, the work of \cite{kocabas2021spec} trains a standalone camera estimation network supervised by the ground-truth camera parameters. However, the ground-truth data is limited and not easy to collect.

We propose a different method for improving the accuracy of cameras by imposing additional constraints in a self-supervised manner. Our main finding is that we can extract multiple RoIs of a human by slightly translating and resizing the original RoI of the human. For each of the RoIs, we then compute the camera projecting the 3D mesh onto the corresponding RoI, which is referred to as a local camera. According to \cite{kissos2020beyond,li2022cliff}, the local cameras of RoIs can be converted to the camera of full image coordinate system. Apparently, all RoIs share the same full camera. We then use the full camera as the intermediate bridge to build pairwise consistency losses between local cameras (see Figure~\ref{fig:figone}).

With the above motivation, we propose a multiple-RoI-based HMR method. At the core of our method is a RoI-aware feature fusion network. It accepts multiple RoIs of a human as input, equipped with a RoI-guided mechanism extracting and fusing features of the multiple RoIs. We obtain two kinds of fusion features: RoI-shared fusion feature and RoI-specific fusion features. The former is decoded to the 3D mesh shared by all RoIs, and the latter are decoded to parameters of local cameras. We then deduce the pairwise camera consistency losses and impose them on the estimated local cameras to regularize the training of the network. Notably, the introducing of the multiple RoIs allow us to encapsulate our network into the contrastive learning framework as RoIs of the same human shall own similar features, while RoIs of different humans shall output dissimilar features. We propose a contrastive loss to enforce this property, which further improves the performance of our method.

In summary, our method is motivated by a simple intuition about the entanglement of mesh and camera. To solve the problem, we propose to extract multiple RoIs, which is novel in this field as most previous approaches are based on a single RoI. 
Our contributions are:
\begin{itemize}
\item We propose a multi-RoI-based HMR method implemented as a RoI-aware feature extraction and fusion network.  
\item We design two loss functions to guide the training of the network, namely a camera consistency loss and a contrastive loss on the basis of the proposed multiple-RoI setting. 
\item Extensive comparisons and ablations validate the designs of our method. 
\end{itemize}

\section{Related Work}
\label{sec:related work}

\textbf{Top-down HMR Methods.} Most approaches recover human mesh in a top-down manner, \ie, cropping the target person from the image and estimating the human mesh of SMPL-based model \cite{loper2015smpl,pavlakos2019expressive, romero2022embodied, osman2020star} in one cropped RoI. There are optimization-based approaches \cite{bogo2016keep,pavlakos2019expressive,fan2021revitalizing}, regression-based approaches \cite{kanazawa2018end, pavlakos2018learning, khirodkar2022occluded, zhang2021body,kolotouros2019learning,wang2018pixel2mesh}, and hybrid approaches \cite{kolotouros2019learning, joo2021exemplar,iqbal2021kama, zhang2021pymaf,li2021hybrik, li2023niki, shetty2023pliks, fang2023learning}. Optimization approaches either fit parameters of a SMPL-based model 
to 2D joints in the input image \cite{bogo2016keep, pavlakos2019expressive}, or fine-tune a pre-trained regression network to match 2D evidences \cite{joo2021exemplar}. Different from optimization-based approaches, regression approaches train a model to extract features from an input image and map the features to a human mesh model, using CNN \cite{pavlakos2018learning, khirodkar2022occluded, zhang2021body, li2022cliff}, GCN \cite{kipf2016semi, choi2020pose2mesh, moon2020i2l}, or Transformer \cite{vaswani2017attention, lin2021mesh, dou2022tore, cho2022cross, lin2022mpt, xue20223d}. Some approaches combine regression and optimization methods. For example, work of \cite{kolotouros2019learning, joo2021exemplar} get an initial prediction through regression-based methods and iteratively optimize the result making it in line with 2D-keypoints reprojection loss. Taking human-kinematics into consideration, work of \cite{li2021hybrik, li2023niki, shetty2023pliks} incorporate Inverse Kinematics Process with the neural network and iteratively update the rotation and location of each joint.

\textbf{HMR with Multiple Inputs.} Considering that HMR is a task with ambiguity, many methods tend to add more auxiliary information at the input end to assist the network to reconstruct the body mesh. Some methods manage to estimate the mesh with the aid of extra inputs such as 2D segmentation or silhouettes of the target human \cite{guan2009estimating, kocabas2021pare, yu2021skeleton2mesh, zhang2020object, yao2019densebody, zhang2020learning} which help the network grasp and understand the human bodies in images with those guidance. Work of \cite{loper2014mosh, AMASS:2019, zanfir2021thundr} try to utilize available sparse 3D markers on surface of the target human before full-body reconstruction and complete the dense human meshes through optimization or interpolation. There are also multi-view methods, by which the ambiguity of HMR is alleviated since multiple view angles and camera parameters are available \cite{sengupta2021probabilistic, li20213d, shin2020multi, pavlakos2022human}. 
A large number of temporal (video-based) methods incorporate auxiliary inputs as well, such as trajectory \cite{yuan2022glamr, goel2023humans}, optical flows \cite{li2022deep} and 3D scene point cloud \cite{zhang2023probabilistic}. There is also egocentric work \cite{liu2023egohmr} using an extra scaled RoI to aid the network to estimate SMPL poses.

\textbf{Approaches Improving Cameras.} There has been much work focusing on the camera projection model since it is the vital bridge between the 2D image and the 3D mesh. Based on \cite{kolotouros2019learning} and \cite{bogo2016keep},  the work of \cite {kissos2020beyond} optimizes the full perspective camera of the original full image for the first time. \cite {kocabas2021spec} tries to estimate camera pitch and yaw angles along with the mesh prediction. \cite{wang2023zolly} introduces a new dataset and copes with the scenario where people are shown up close in the image, taking the distortion of perspective projection into consideration. CLIFF\cite{li2022cliff} digs deeper into the full-image reprojection and uses boundingbox information in order to guide the network towards the accurate full camera. 
We, in this paper, incorporate the theory of full-image projection in \cite{li2022cliff, kissos2020beyond} and model the pairwise relations between cameras estimated from different RoIs of the same person. 

\begin{figure*}[!t]
\centering
\includegraphics[width=1.0\textwidth]{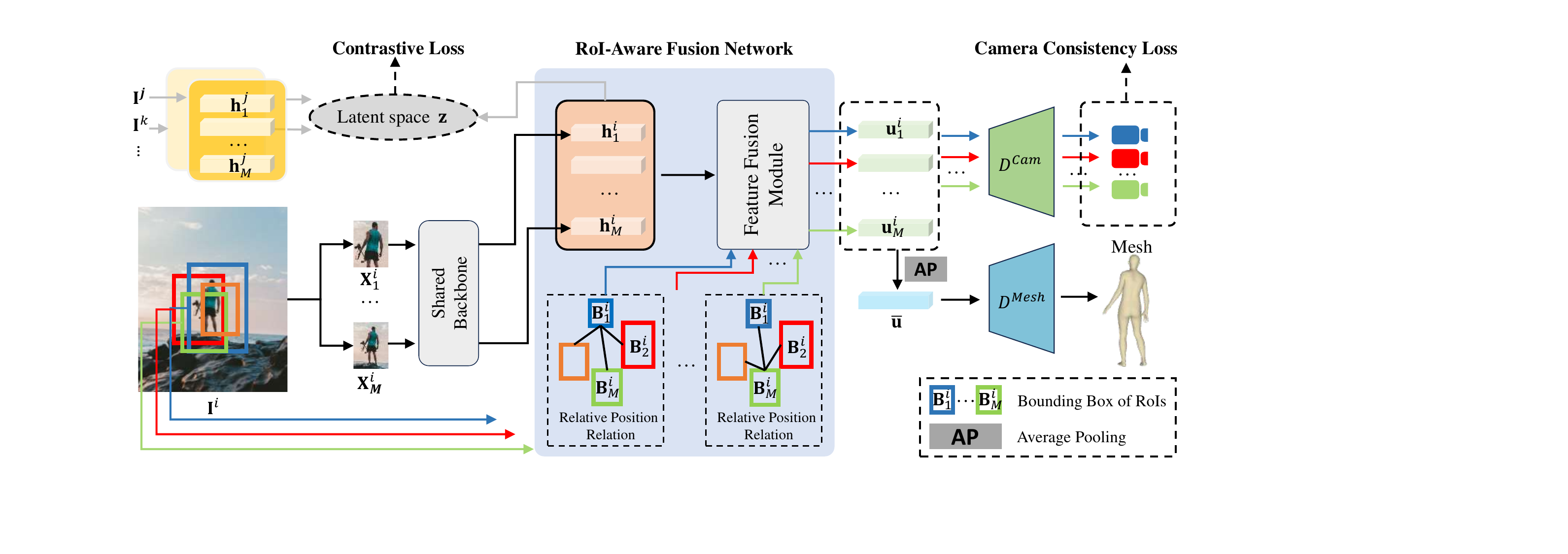}
\caption{\textbf{Overview of our method.} Given an image, we extract multiple RoIs of a human, and use a RoI-aware feature fusion network to estimate the 3D mesh of the human together with cameras. We use a camera consistency loss and a contrastive loss to supervise the training of the network.} 
\label{fig:overview}
\end{figure*}

\section{Method}
\label{sec:main}

Figure~\ref{fig:overview} provides the overview of our method. Given a full image $\mathbf{I}^i$, we extract $M$ RoIs $\{\mathbf{X}^i_m\}_{m=1}^M$ of a person in the image by different boundingboxes $\{\mathbf{B}^i_m\}_{m=1}^M$, and use a shared backbone network to extract features $\{\mathbf{h}^i_m\}_{m=1}^M$ from the RoIs. Then, we propose a RoI-aware fusion network to fuse $\{\mathbf{h}_m^i\}_{m=1}^M$ to obtain RoI-specific fusion features $\{\mathbf{u}_m^i\}_{m=1}^M$ and a RoI-shared fusion feature $\bar{\mathbf{u}}$. Each RoI-specific feature is individually decoded by $D^{cam}$ to a local camera, obtaining $M$ local cameras to which we apply camera consistency loss. The RoI-shared feature is decoded by $D^{mesh}$ to the target 3D mesh. We also extract features from RoIs of other objects (\eg, in $\mathbf{I}^j$, $\mathbf{I}^k$) and project all of them into the latent space of $\mathbf{z}$, and finally apply a contrastive loss in the $\mathbf{z}$-space.

Formally, the regression task in this paper is formulated as:
\begin{equation}
\theta,\beta,\{\mathbf{C}_m\}_{m=1}^{M} = f(\{\mathbf{X}_m\}_{m=1}^M, \{\mathbf{B}_m\}_{m=1}^M),
\end{equation}
where $\mathbf{\theta}\in\mathbb{R}^{24\times 3}$ determines the pose of the SMPL mesh, $\mathbf{\beta}\in\mathbb{R}^{10}$ determines the shape of the SMPL mesh, and $\mathbf{C_m} = (s_m,t_{x_m},t_{y_m})$ contains scale $s_m$ and translation parameters $(t_{x_m},t_{y_m})$ determining a weak-perspective camera that projects the predicted 3D mesh onto the 2D RoI plane. $\mathbf{B}_m = (c_{x_m}, c_{y_m}, b_m) $, where $(c_{x_m},c_{y_m})$ is the location of the boundingbox in the full image, and $b_m$ is the width of the boundingbox.

\subsection{RoI-aware Feature Fusion Network}
\label{subsec: RAF}
To be specific, given $\{\mathbf{X}_m\}_{m=1}^{M}$ (the superscript $i$ used in Figure~\ref{fig:overview} is dropped for simplicity), we use a shared encoder $E$ to extract features from RoIs, \ie, $\mathbf{h}_m=E(\mathbf{X}_m)$ for $m \in [1, M]$. 
The encoder $E$ can be ResNet50 \cite{he2016identity} 
or HRNet48 \cite{sun2019deep} 
as employed in previous approaches \cite{kanazawa2018end, li2022cliff, black2023bedlam, cho2022cross}. 
After that, we design a RoI-aware fusion network to fuse $\{\mathbf{h}_m\}_{m=1}^{M}$, obtaining RoI-specific fusion features $\{\mathbf{u}_m\}_{m=1}^{M}$. 
Then, we simply average the RoI-specific features by AP (Average Pooling) to obtain the RoI-shared feature $\bar{\mathbf{u}}$. 

\begin{figure}[!t]
\centering
\includegraphics[width=1.0\columnwidth]{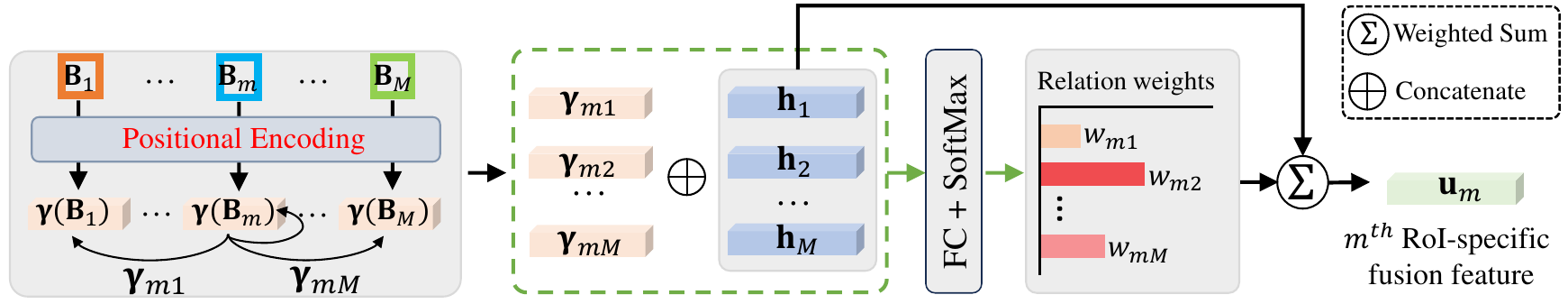}
\caption{\textbf{RoI-aware fusion}. To obtain $\mathbf{u}_m$, we consider the relative relation of other boundingboxes to the $m^{th}$ boundingbox. We perform positional encoding to all the boundingboxes and then compute relative position relation $\gamma_{m*}$ (where $*$ is a number in $[1,M]$). We then concatenate $\gamma_{m*}$ and the corresponding feature $\mathbf{h}_*$ to compute weight $w_{m*}$. Finally, $\mathbf{u}_m$ is the weighted sum of $\{\mathbf{h}_m\}_{m=1}^M$ with $w_{m*}$ as the weights.}   
\label{fig:crop-aware-fusion}
\end{figure}

The core of our network is the feature fusion module. To begin with, the feature $\mathbf{h}_m$ only contains the information about the $m^{th}$ RoI. Since different RoIs contain different visual details about the target person, we fuse all features $\{\mathbf{h}_m\}_{m=1}^{M}$ together for reasoning about the mesh and cameras. 
Our fusion method, as illustrated in Figure \ref{fig:crop-aware-fusion}, leverages the boundingbox information $\{\mathbf{B}_m\}_{m=1}^{M}$, 
by which we compute the \textit{relative} position relation between the boundingboxes to align the features of different RoIs.  
Specifically, the relative position relation is simply computed as the pairwise difference between boundingboxes after positional encoding:
\begin{equation}
\gamma_{mn} = \gamma(\mathbf{B}_m)-\gamma(\mathbf{B}_n),
\end{equation}
where $\gamma(\cdot)$ is the position encoding function \cite{mildenhall2021nerf, vaswani2017attention}:
\begin{equation}
\small
\gamma(p) = (p, {\rm sin}(\pi p),{\rm cos}(\pi p), 
\cdots, {\rm sin}(2^{L}\pi p),{\rm cos}(2^{L}\pi p),
\end{equation}
which is applied to each of the three variables of $\mathbf{B}_m$ (or $\mathbf{B}_n$). We set $L = 32$ in this paper. Then, taking the $m^{th}$ RoI as example, the way to compute the fused feature $\mathbf{u}_m$ is:
\begin{equation}
\begin{aligned}
\mathbf{u}_m &= \sum_{n=1}^{M} w_{mn} \mathbf{h}_n, \\
\{w_{mn}\}_{n=1}^M &= {\rm Softmax}({\rm Linear}(\{\textbf{F}_{mn}\}_{n=1}^M)) \\
\textbf{F}_{mn} &= {\rm Concat}(\mathbf{h}_n,\gamma_{mn}),
\end{aligned}
\end{equation}
where 
$w_{mn}\in[0,1]$ is a scalar used to fuse features of multiple RoIs. To compute $w_{mn}$, we first concatenate the feature $\mathbf{h}_n$ and the relative position relation $\gamma_{mn}$, and then send the concatenated feature to a linear layer to obtain a scalar, which is finally converted to $w_{mn}$ by a Softmax function. 
Eventually, we use a $D^{cam}$, which is composed of FC layers with residual connections as adopted in \cite{kanazawa2018end}, to compute the local camera $\mathbf{C}_m$ from the feature $\mathbf{u}_m$:
\begin{equation}
\mathbf{C}_m = D^{cam}(\mathbf{u}_m).
\label{eq:decoder-cam}
\end{equation}
We employ $D^{mesh}$ similar to $D^{cam}$ to map the averaged feature $\bar{\mathbf{u}}$ to 3D mesh:
\begin{equation}
\theta,\beta = D^{mesh}(\bar{\mathbf{u}}).
\label{eq:decoder-mesh}
\end{equation}

\begin{figure*}[!t]
\centering
\includegraphics[width=0.8\textwidth]{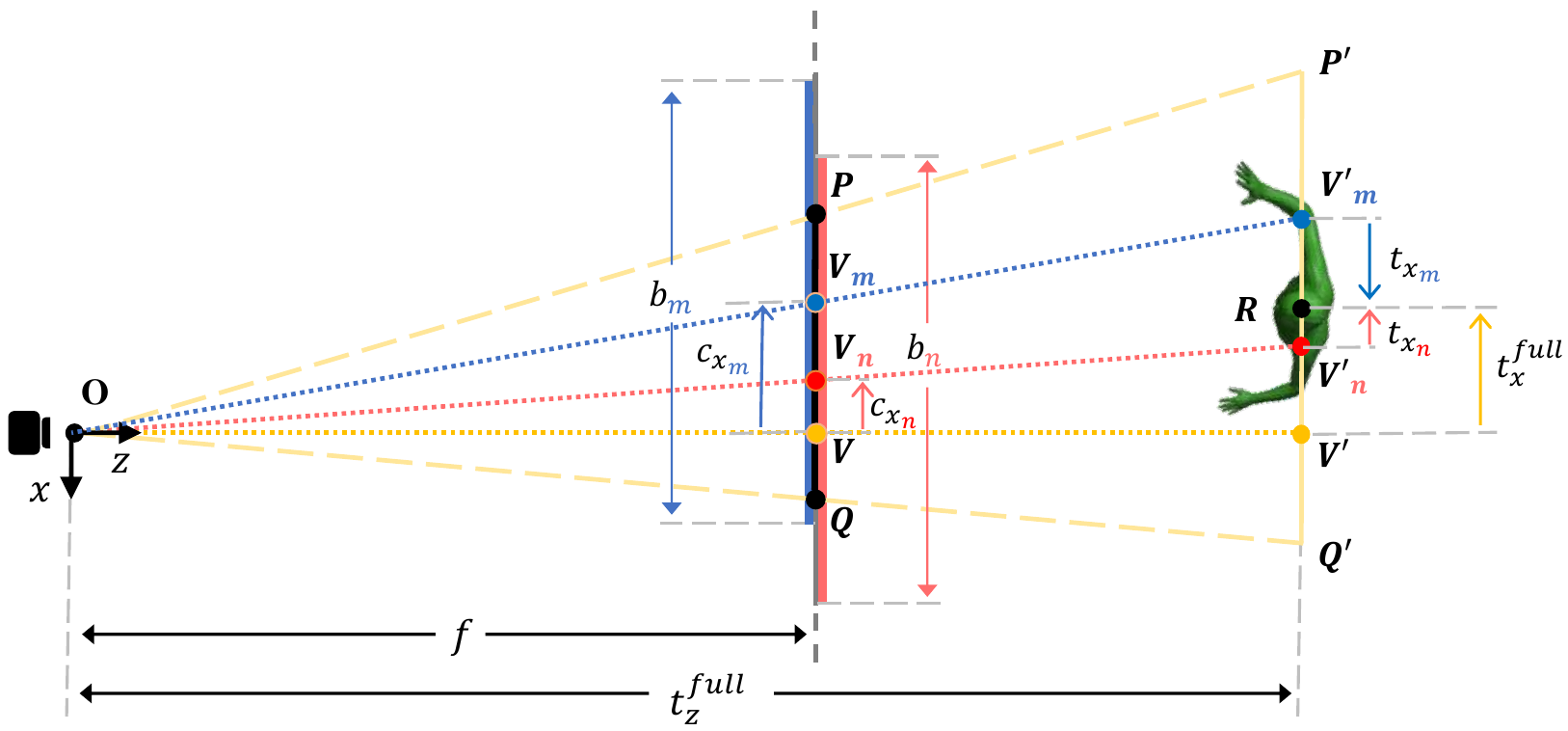}
\caption{{Conversion between local and full cameras in bird's eye view.} }
\label{fig:cam_los}
\end{figure*}

\subsection{Camera Consistency Loss}
To build camera consistency loss, local cameras are converted to the coordinate system of full camera according to \cite{kissos2020beyond,li2022cliff}. Formally, let $\mathbf{C}_m=(s_m,t_{x_m},t_{y_m})$ and $\mathbf{C}_n=(s_n,t_{x_n},t_{y_n})$ be two local cameras estimated from two RoIs cropped by boundingboxes $\mathbf{B}_m=(c_{x_m}, c_{y_m}, b_m)$ and $\mathbf{B}_n=(c_{x_n}, c_{y_n}, b_n)$, respectively. Let $\mathbf{C}_{full}=(t_{x}^{full},t_{y}^{full},t_{z}^{full})$ be the parameters of the full camera. The focal length of the full camera is denoted as $f$. We refer to Figure~\ref{fig:cam_los} to interpret the conversion between these variables.

Figure~\ref{fig:cam_los} shows that a 3D mesh at distance $t_z^{full}$ from the camera $O$ is projected onto the image plane in a focal length of $f$. We assume the 3D human mesh is bounded in a 2m-2m-2m box. From the bird's eye view, we use $P'Q'$ to denote the valid region occupied by the 3D mesh, and the length of $P'Q'$ is 2m, which is projected to $PQ$ on the image plane. The blue line on the image plane indicates the RoI $\mathbf{B}_m$, the length of which is $b_m$, \ie, the width of the bounding box. 
Since $\bigtriangleup OPQ$ and $\bigtriangleup OP'Q'$ are similar, we have:
\begin{equation}
\footnotesize
    \frac{PQ}{P'Q'}=\frac{f}{t_z^{full}}, \quad\ie,\quad\frac{b_m\cdot s_m}{2}=\frac{f}{t_z^{full}},
\end{equation}
where $b_m\cdot s_m$ is the length of $PQ$. 
And we get,
\begin{equation}
\footnotesize
    t_z^{full} = \frac{2\cdot f}{b_m\cdot s_m}.
    \label{eq:define_of_tzfull}
\end{equation}
On the other hand, let $V_m$ be the center of $\mathbf{B}_m$ on the image plane. The distance from $V$ (the image center) to $V_m$ is $c_{x_m}$. Let $V'_m$ be the point on the mesh plane corresponding to $V_m$, and $R$ be the center of mesh. The distance from $V'_m$ to $R$ is the local camera translation $t_{x_m}$. The distance from $V'$ to $R$ is the full camera translation $t_x^{full}$. Since $\bigtriangleup OVV_m$ and $\bigtriangleup OV'V'_m$ are similar, we have:
\begin{equation}
\footnotesize
    \frac{VV_m}{V'V'_m}=\frac{f}{t_z^{full}},
    \quad\ie,\quad \frac{c_{x_m}}{t_x^{full}-t_{x_m}}=\frac{f}{t_z^{full}}.
    \label{eq:delta_x}
\end{equation}
Combining Eq.~\ref{eq:define_of_tzfull} with Eq.~\ref{eq:delta_x}, we get
\begin{equation}
    t_x^{full} = t_{x_m} + \frac{2\cdot c_{x_m}}{b_m \cdot s_m}
    \label{eq:define_of_txfull}
\end{equation}
The above is the relation between local translation $t_{x_m}$ and global translation $t_x^{full}$. 
Similarly, the relation along the $y$ axis is:
\begin{equation}
\footnotesize
    t_y^{full} = t_{y_m} + \frac{2\cdot c_{y_m}}{b_m \cdot s_m}.
    \label{eq:define_of_tyfull}
\end{equation}
From Eq. \ref{eq:define_of_txfull}, \ref{eq:define_of_tyfull} and \ref{eq:define_of_tzfull}, we convert the local camera $\mathbf{C}_m$ to the full camera $\mathbf{C}_{full}$ by:
\begin{equation}
\footnotesize
    t_x^{full} = t_{x_{m}}+\frac{2 \cdot c_{x_{m}}}{b_{m}\cdot{s_{m}}},\quad
		t_y^{full} = t_{y_{m}}+\frac{2 \cdot c_{y_{m}}}{b_{m}\cdot{s_{m}}},\quad
		t_z^{full} = \frac{2 \cdot f}{b_{m}\cdot{s_{m}}}.
   \label{eq:cliff_cam_mtofull}
\end{equation}
Similarly, we can convert local camera $\mathbf{C}_n$ to the full camera $\mathbf{C}_{full}$ by:
\begin{equation}
\footnotesize
		t_x^{full} = t_{x_{n}}+\frac{2 \cdot c_{x_{n}}}{b_{n}\cdot{s_{n}}},\quad
		t_y^{full} = t_{y_{n}}+\frac{2 \cdot c_{y_{n}}}{b_{n}\cdot{s_{n}}},\quad
		t_z^{full} = \frac{2 \cdot f}{b_{n}\cdot{s_{n}}}.
    \label{eq:cliff_cam_ntofull}
\end{equation}
Combining Eq.~\ref{eq:cliff_cam_mtofull} and \ref{eq:cliff_cam_ntofull}, we establish the following relations between parameters of local cameras:
\begin{equation}
\left\{
\begin{array}{cl}
&  t_{x_m} + \frac{{2 \cdot {c_{{x_m}}}}}{{{b_m} \cdot {s_m}}} = t_{x_n} + \frac{{2 \cdot {c_{{x_n}}}}}{{{b_n} \cdot {s_n}}} \\
&  t_{y_m} + \frac{{2 \cdot {c_{{y_m}}}}}{{{b_m} \cdot {s_m}}} = t_{y_n} +  \frac{{2 \cdot {c_{{y_n}}}}}{{{b_n} \cdot {s_n}}}  \\
& {b_m} \cdot {s_m} = {b_n} \cdot {s_n}
\end{array}
\right.
\label{eq:camloss1}
\end{equation}
We define
\begin{equation}
\footnotesize
\left\{
\begin{array}{cl}
\mathcal{L}_x(m,n) &= \left \| \left(t_{{x_m}} + \frac{{2 \cdot {c_{{x_m}}}}}{{{b_m} \cdot {s_m}}}\right) - \left(t_{{x_n}} + \frac{{2 \cdot {c_{{x_n}}}}}{{{b_n} \cdot {s_n}}}\right)\right \| ^2_2\\
\mathcal{L}_y(m,n) &= \left \| \left(t_{{y_m}} + \frac{{2 \cdot {c_{{y_m}}}}}{{{b_m} \cdot {s_m}}}\right) - \left(t_{{y_n}}  + \frac{{2 \cdot {c_{{y_n}}}}}{{{b_n} \cdot {s_n}}}\right)\right \|^2_2\\
\mathcal{L}_s(m,n) &= \left \|b_m\cdot s_m - b_n\cdot s_n\right \|_2^2 \\
\end{array}
\right.
\label{eq:camloss2}
\end{equation}
Finally, the local camera consistency loss is defined as:
\begin{equation}
\small
\mathcal{L}_{cam} = \sum_{m,n}^{M}\lambda _x\mathcal{L}_x(m,n) + \lambda _y \mathcal{L}_y(m,n) + \lambda _s\mathcal{L}_s(m,n),
\label{eq:camera-consistency-loss}
\end{equation}
where $\lambda_x$, $\lambda_y$ and $\lambda_s$ are weights of the three regularization terms, which are 0.1, 0.1 and 0.0001, respectively. 

\subsection{Contrastive Loss}
\label{sec: contrastive module}
An extra benefit of using multiple RoIs as input is that we can apply a contrastive loss as another regularization term besides the camera consistency loss.  
At training, we have access to RoIs of different persons. It is natural to require extracting similar features from RoIs of the same person. While for RoIs of different persons, different features should be extracted. The contrastive learning of~\cite{chen2020simclr} can be adapted to fulfill this purpose. 

\begin{figure*}[!t]
\centering
\includegraphics[width=1.0\textwidth]{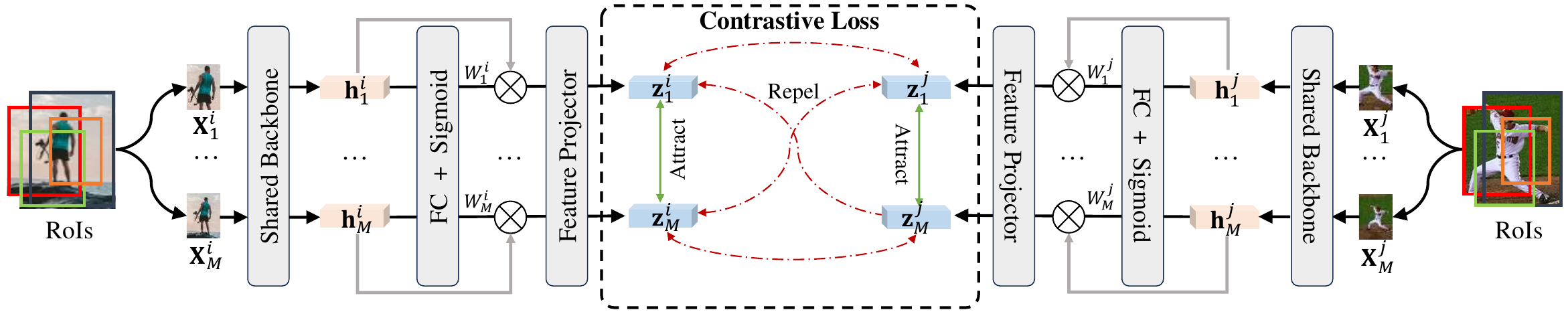}
\caption{\textbf{Contrastive Loss.} Taking RoIs $\{\mathbf{X}^i_m|m\in[1,M]\}$ of object $i$ and RoIs $\{\mathbf{X}^j_m|m\in[1,M]\}$ of object $j$ as example, features $\{\mathbf{h}^i_m|m\in[1,M]\}$ and $\{\mathbf{h}^j_m|m\in[1,M]\}$ are first extracted by the shared backbone $E$ from the RoIs, respectively. Then the features are further projected into the latent space $\mathbf{z}$, obtaining $\{\mathbf{z}^i_m|m\in[1,M]\}$ and $\{\mathbf{z}^j_m|m\in[1,M]\}$. The latent features from the same object attract each other, while latent features from different objects repel each other.}
\label{fig:contrastive}
\end{figure*}

Let $\{\mathbf{X}^i_m|m\in[1,M]\}$ be RoIs of object $i$ with $i\in[1,N]$ where $N$ is the number of objects in a training batch. We first extract features from all the RoIs, obtaining $\{\mathbf{h}^i_m|i\in[1,N],m\in[1,M]\}$. Then we further project the features into a latent space $\mathbf{z}$, obtaining latent features  $\{\mathbf{z}^i_m|i\in[1,N],m\in[1,M]\}$. Figure~\ref{fig:contrastive} illustrates the the mapping process from $\mathbf{X}$ to $\mathbf{z}$. The contrastive loss is defined on all the latent features:
\begin{equation}
	\footnotesize
		{\mathcal{L}_{cont}} = \sum\limits_{i = 1}^{N}\sum_{m=1}^{M} {\frac{-1}{M-1}} \sum\limits_{n=1,n\neq m}^{M} \log {\frac{\exp (\mathbf{z}_m^i\cdot \mathbf{z}_n^i/\tau)}{\sum\limits_{i' = 1,i'\neq i}^{N} \sum\limits_{m'=1}^{M} \exp (\mathbf{z}_m^i\cdot \mathbf{z}_{m'}^{i'}/\tau)}},
		\label{eq:contrastive-loss}
\end{equation}
where 
$\tau=0.5$. 
The numerator/denominator aims at (1) minimizing cosine distance between features $\mathbf{z}_m^i$ and $\mathbf{z}_n^i$ from the same object $i$, and (2) maximizing distance between features $\mathbf{z}_m^i$ and $\mathbf{z}_{m'}^{i'}$ from different objects $i$ and $i'$.

\subsection{Total Training Loss}
\label{sec: losses}
Besides the the camera consistency loss in Eq.~\ref{eq:camera-consistency-loss} and contrastive loss in Eq.~\ref{eq:contrastive-loss}, we also adopt the typical losses using GT mesh and 2D joints as supervision: 
\begin{equation}
\footnotesize
\begin{aligned}
\mathcal{L}_{smpl} = \left\|\Theta-\hat{\Theta} \right \|, &\quad
\mathcal{L}_{vert} = \left \|V^{3D}-\hat{V}^{3D}\right \|,  \\
\mathcal{L}_{3D} = \left\|J^{3D}-\hat J^{3D}\right \|^2_2 , &\quad
\mathcal{L}_{2D} = \sum_m^{M}\left \|J^{2D}_m-\hat J^{2D}\right \|^2_2,
\end{aligned}
\end{equation}
where $\Theta=(\theta,\beta)$ denotes estimated SMPL parameters and $\hat{\Theta}$ is the ground truth (GT), $V^{3D}$ indicates 3D vertices of human mesh with $\hat{V}^{3D}$ as GT, and $J^{3D}$ denotes the 3D joints of the human with $\hat J^{3D}$ as GT. For the 2D reprojection loss, $J^{2D}_m$ is obtained by projecting $J^{3D}$ from 3D to 2D with the full camera deduced from local camera $\mathbf{C}_m$. 
Following \cite{li2022cliff}, the projected joints are compared with the GT 2D joints $\hat J^{2D}$ in the full image. The total loss function is:
\begin{equation}
\footnotesize
\mathcal{L}_{total} = \lambda _{cam} \mathcal{L}_{cam}+\lambda _{cont} \mathcal{L}_{cont}+
\lambda _{smpl}\mathcal{L}_{smpl}+\lambda _{vert} \mathcal{L}_{vert} +\lambda _{3D}\mathcal{L}_{3D}+\lambda _{2D} \mathcal{L}_{2D},
\label{eq:loss_overall}
\end{equation}
where $\lambda_{*}$ are weights for each loss component and we set them following SPIN~\cite{kolotouros2019learning} except $\lambda_{cont}$ and $\lambda_{cam}$ which are 0.1 and 1 respectively.

\subsection{Extraction of RoIs}
Given an image, we use methods of  \cite{he2017mask, jocher2022ultralytics} to detect boundingboxes of human. Let $\mathbf{B}=(c_x,c_y,b)$ be a boundingbox, we slightly resize/translate the the boundingbox to select multiple RoIs of the human. One can randomly generate the resizing factors or translation offsets. However, experiments (see Supp.) show that fixing these parameters during training gives better results. Specifically, the offset along the $x$ and $y$ axes includes $\{(0.1b,0),(-0.1b,0),(0,0.1b),(0,-0.1b)\}$, and the corresponding resizing factors are $\{1.5,1.25,0.8,0.65\}$. Together with the original boundingbox, we totally extract $M=5$ RoIs for a person from the full image. More detailed illustrations are provided in supplemental material.

\section{Experiments}
\label{sec:exp}
\subsection{Datasets and Metrics}
\label{subsec:data}
To conduct fair comparison between our method and SOTA methods, we follow the dataset setting used in SOTA works \cite{zhang2021pymaf, cho2022cross, li2021hybrik, black2023bedlam, kocabas2021pare, cheng2023bopr}. Specifically, we train our method on a mixture of four datasets including Human3.6M~\cite{ionescu2013human3}, MPI-INF-3DHP \cite{mehta2017monocular}, COCO~\cite{lin2014microsoft}, and MPII~\cite{andriluka20142d}. 

As for evaluation, 
we use the test sets of 3DPW~\cite{von2018recovering} and Human3.6M \cite{ionescu2013human3}. Following prior works, we finetune our model on 3DPW train set when evaluating on its test set.
\footnote[1]{The authors Yongwei Nie and Changzhen Liu signed the license and produced all the experimental results in this paper. Meta did not have access to the datasets.
}

We use MPJPE (Mean Per Joint Position Error \cite{ionescu2013human3}), PA-MPJPE (Procrustes-Aligned MPJPE \cite{zhou2018monocap}) and PVE (the mean Euclidean distance between mesh vertices) as the evaluation metrics. 

\subsection{Implementation Details}
\label{subsec: implement}
We implement our method using PyTorch. For the shared backbone, we use ResNet-50 \cite{he2016identity} extracting features of $d=2048$ dimensions and HRNet-W48 \cite{sun2019deep} extracting features of $d=720$ dimensions, and refer to our methods with these backbones as Ours$^{R50}$ and Ours$^{H48}$, respectively. Following \cite{black2023bedlam, wang2023zolly}, the adopted backbones of ResNet-50 and HRNet-W48 are pre-trained on COCO~\cite{lin2014microsoft} for 2D pose estimation. 
We train our models with a learning rate of 1e-4 and 5e-5 for ResNet and HRNet backbones respectively, both scheduled by an Adam optimizer with $\beta_1=0.9$ and $\beta_2=0.999$. The batchsize for Ours$^{R50}$ is $48$ and that for Ours$^{H48}$ is $20$. Training with ResNet-50 takes 25 epochs for 1 day and training with HRNet-W48 takes 15 epochs for 2 days on NVIDIA RTX 3090. When finetuning on 3DPW, we fix the learning rate at 1e-5 (for both backbones) to train our models for another 5 epochs. By default, we use $M=5$ RoIs.

\begin{table}[!t]\scriptsize
\caption{\textbf{Quantitative comparison with SOTA methods.} $R50$ (or $R34$) denotes using ResNet \cite{he2016identity} as backbone. $H48$ (or $H32$, $H64$) denotes using HRNet \cite{sun2019deep}. Note that we present the result of Zolly$^{H48}$ trained without synthetic distorted data for fairness, as reported in their paper.}
	\centering
	\begin{tabular*}{\columnwidth}{p{3.4cm}<{\raggedright}p{1.6cm}<{\centering}p{1.6cm}<{\centering}p{1.6cm}<{\centering}|p{1.6cm}<{\centering}p{1.7cm}<{\centering}}
			\toprule
			\multirow{2}{*}{Method}& \multicolumn{3}{c|}{3DPW} & \multicolumn{2}{c}{Human3.6M} \\
			\noalign{\smallskip}
			\cline{2-6}
			\noalign{\smallskip}
			& MPJPE & PA-MPJPE &PVE &MPJPE & PA-MPJPE \\
			\midrule
			FastMETRO$^{H64}$~\cite{cho2022cross}'22 & 73.5 & 44.6 & 84.1 & 52.2 & 33.7\\
			CLIFF$^{H48}$~\cite{li2022cliff}'22 &69.0 & 43.0 & 81.2 & 47.1 & 32.7 \\
			BoPR$^{H48}$~\cite{cheng2023bopr}'23 & 65.4 & 42.5 & 80.8 & $-$ & $-$\\
			ReFit$^{H48}$~\cite{wang2023refit}'23 &65.8 & 41.0 & $-$ & 48.4 & 32.2\\	
			PyMAF-X$^{H48}$~\cite{zhang2023pymaf}'23 &74.2 & 45.3 & 87.0 & 54.2 & 37.2\\
			NIKI$^{H48}$~\cite{li2023niki}'23 & 71.3 & 40.6 & 86.6 & $-$ & $-$\\
			\midrule
			Ours$^{H48}$ &\textbf{64.1} & \textbf{40.4} &78.6 & \textbf{42.2} & \textbf{30.7}\\
			
			\bottomrule
		\end{tabular*}

	\label{tab:main_results_}
\end{table}

\begin{figure*}[!t]
\centering
\includegraphics[width=0.104\textwidth]{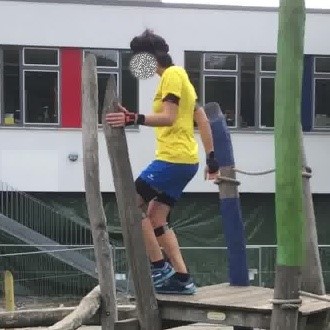}
\includegraphics[width=0.104\textwidth]{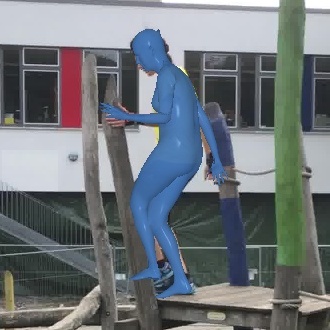}
\includegraphics[width=0.104\textwidth]{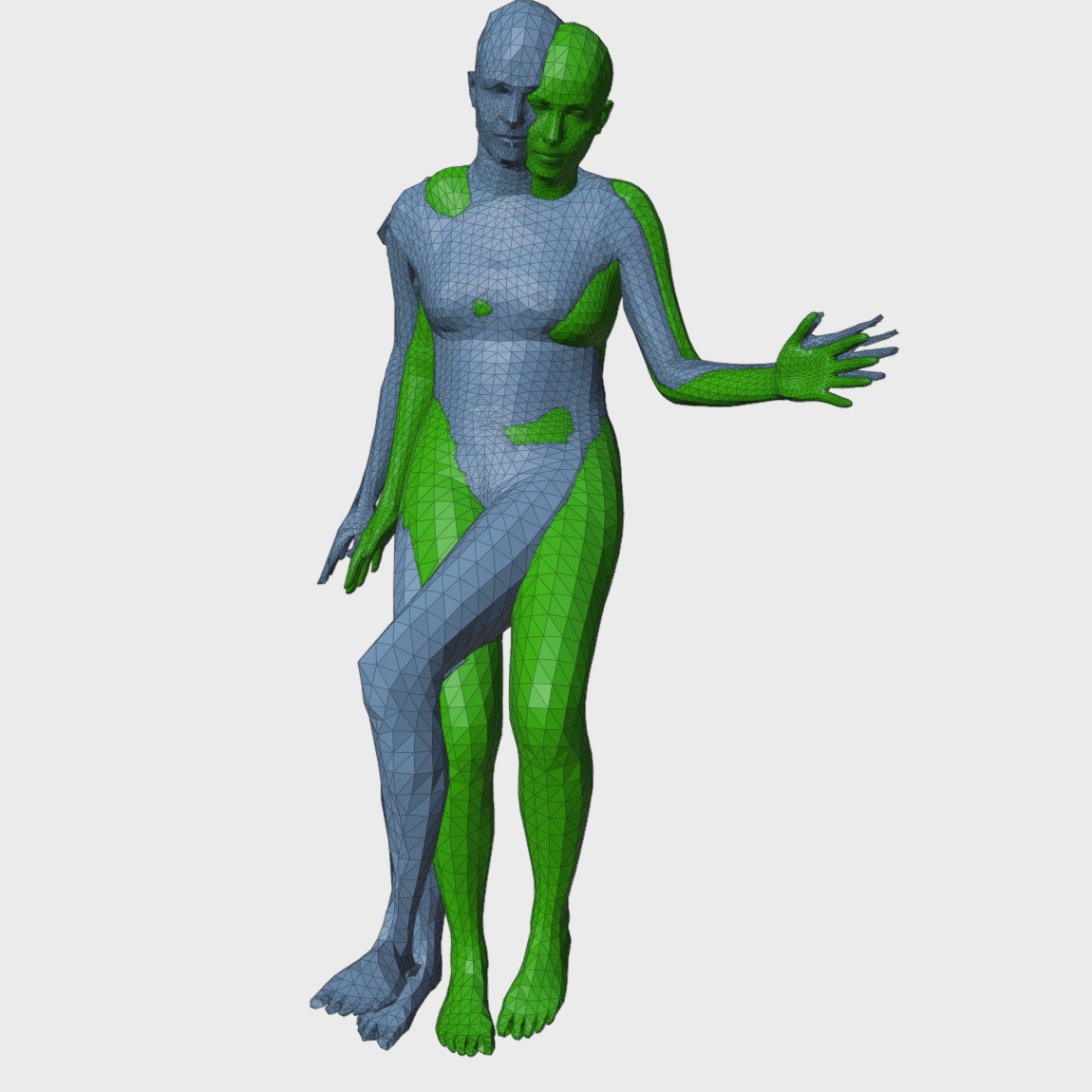}
\includegraphics[width=0.104\textwidth]{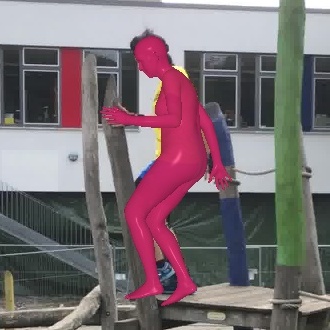}
\includegraphics[width=0.104\textwidth]{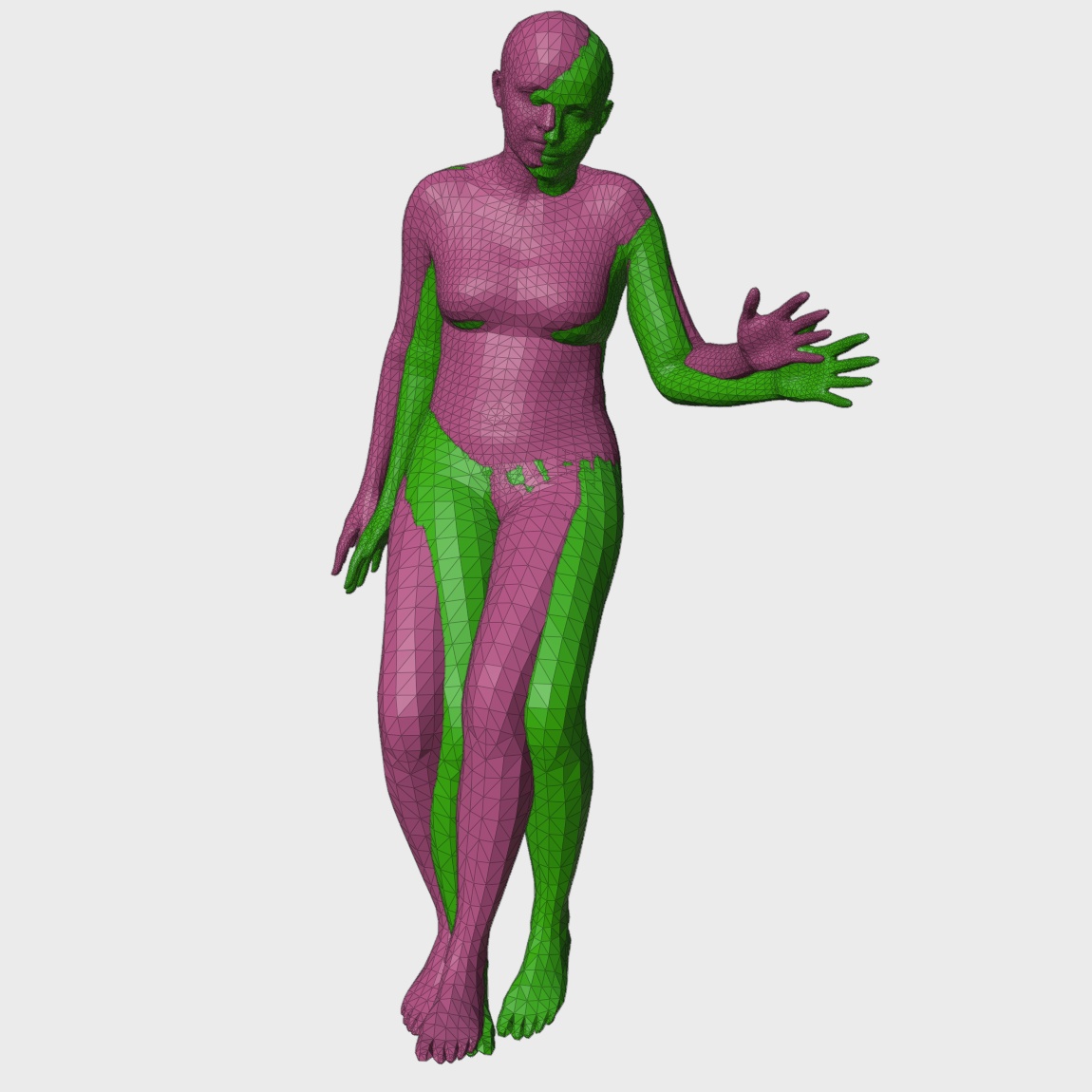}
\includegraphics[width=0.104\textwidth]{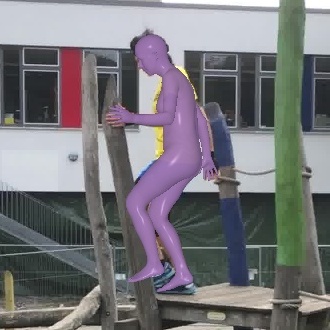}
\includegraphics[width=0.104\textwidth]{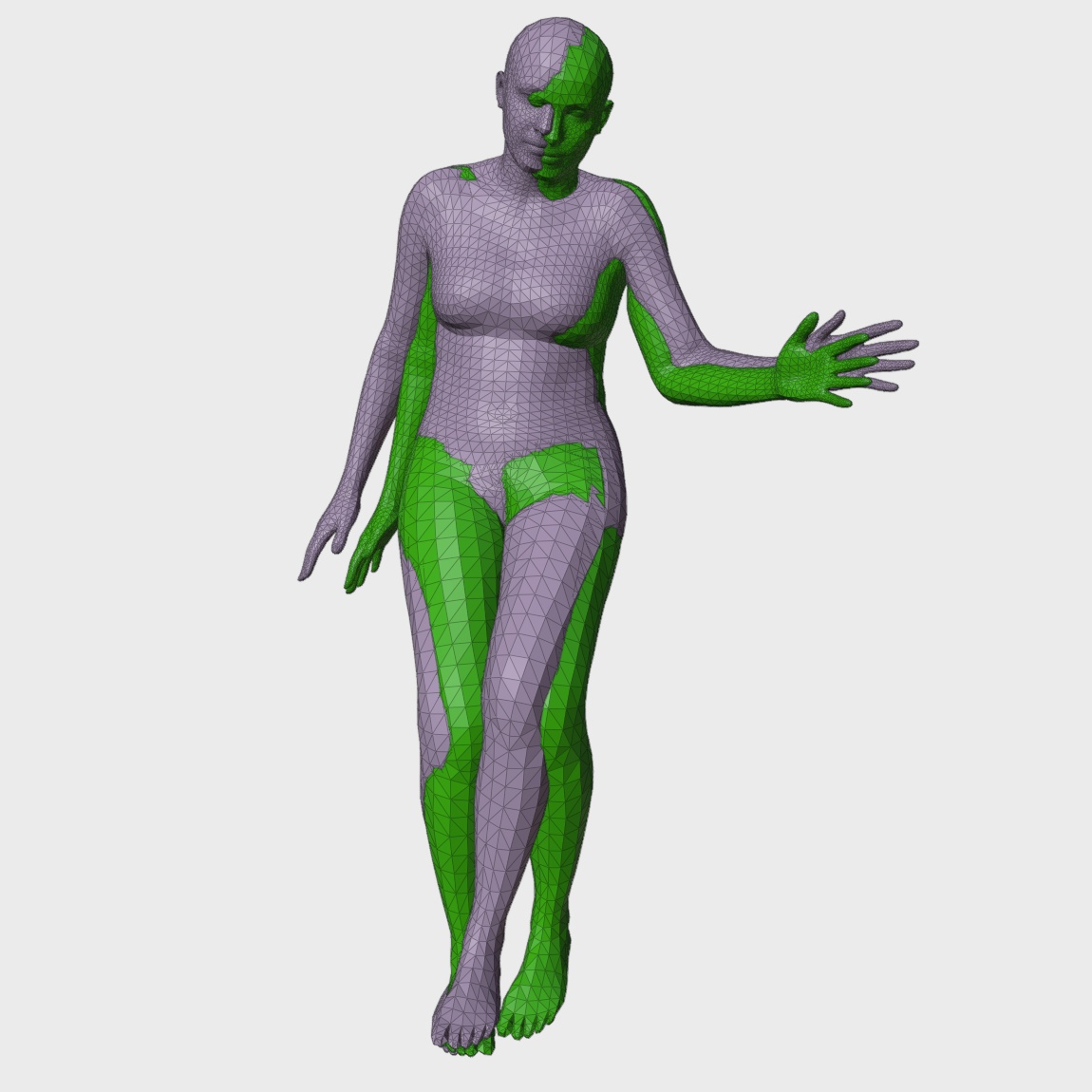}
\includegraphics[width=0.104\textwidth]{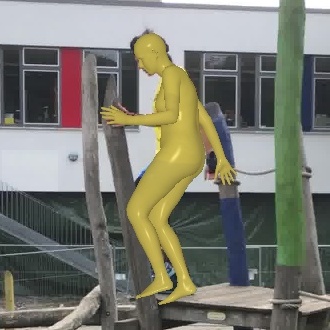}
\includegraphics[width=0.104\textwidth]{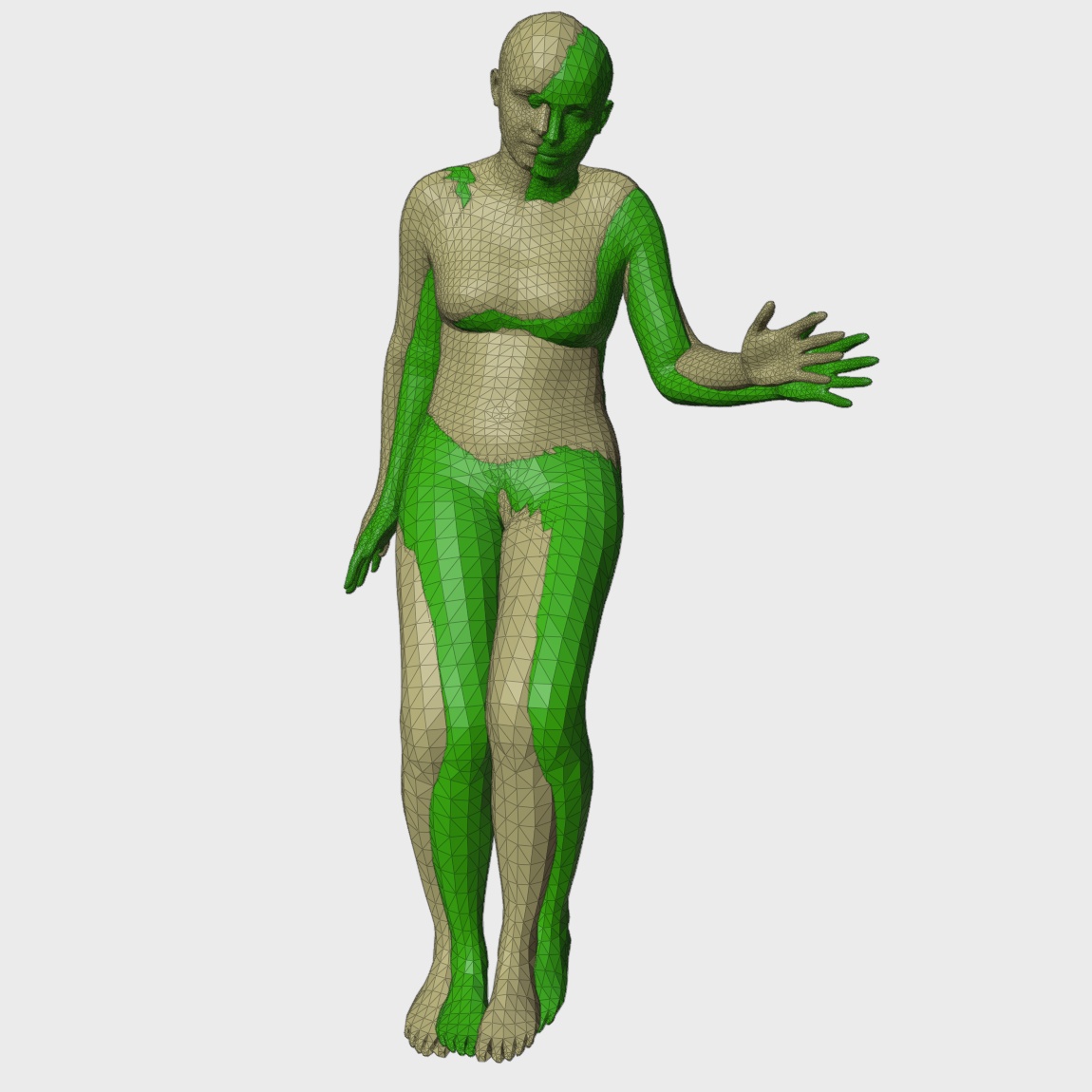} \\
\includegraphics[width=0.104\textwidth]{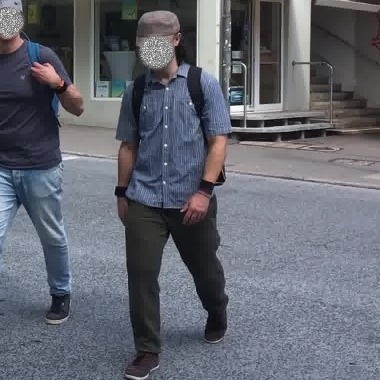}
\includegraphics[width=0.104\textwidth]{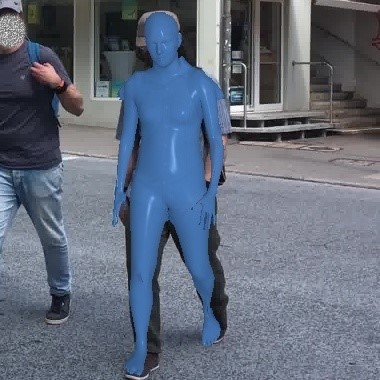}
\includegraphics[width=0.104\textwidth]{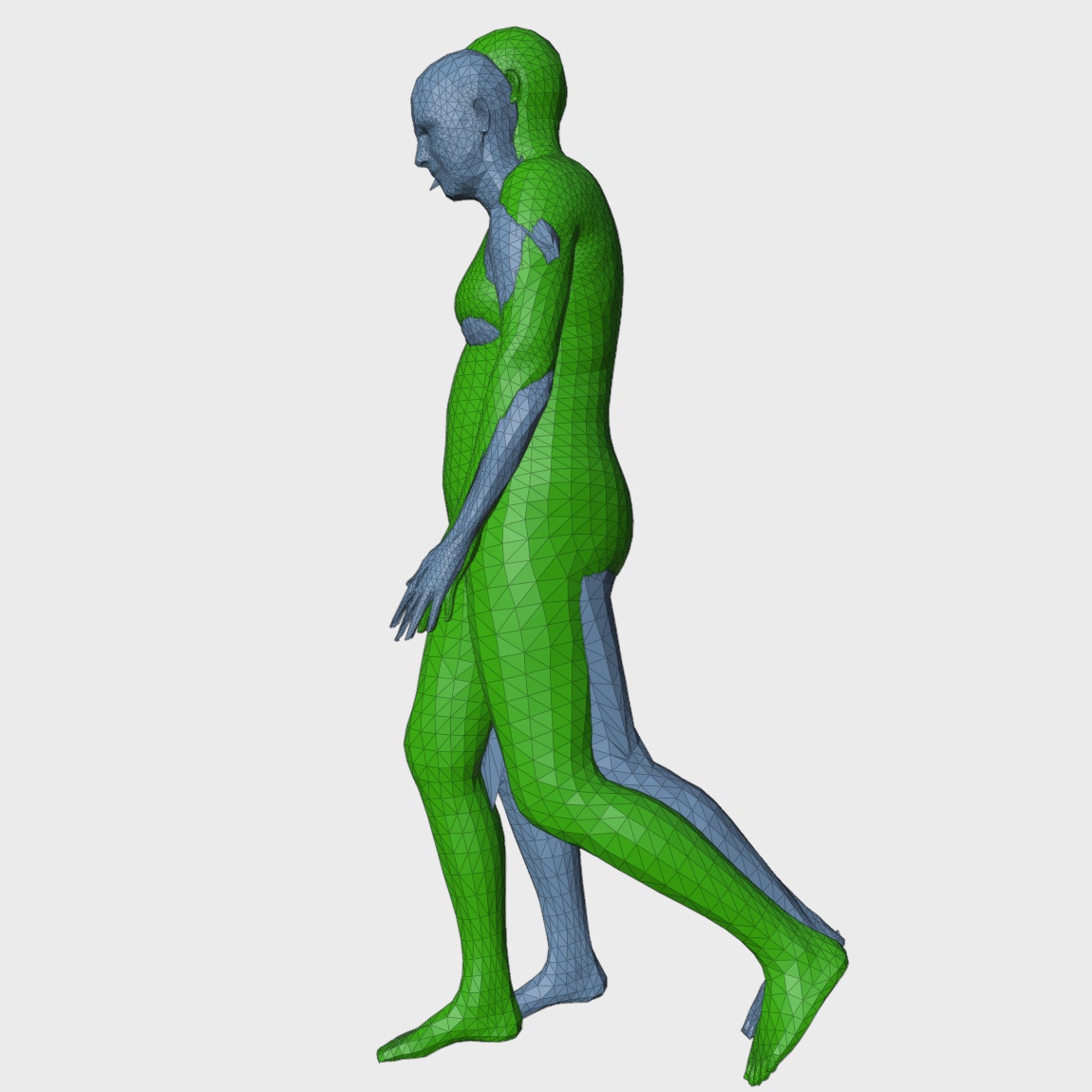}
\includegraphics[width=0.104\textwidth]{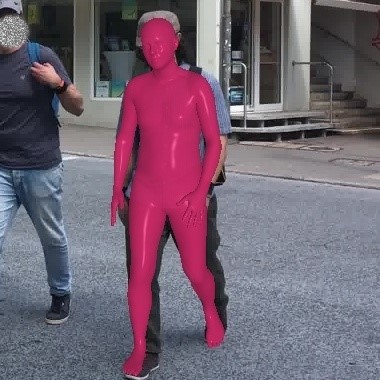}
\includegraphics[width=0.104\textwidth]{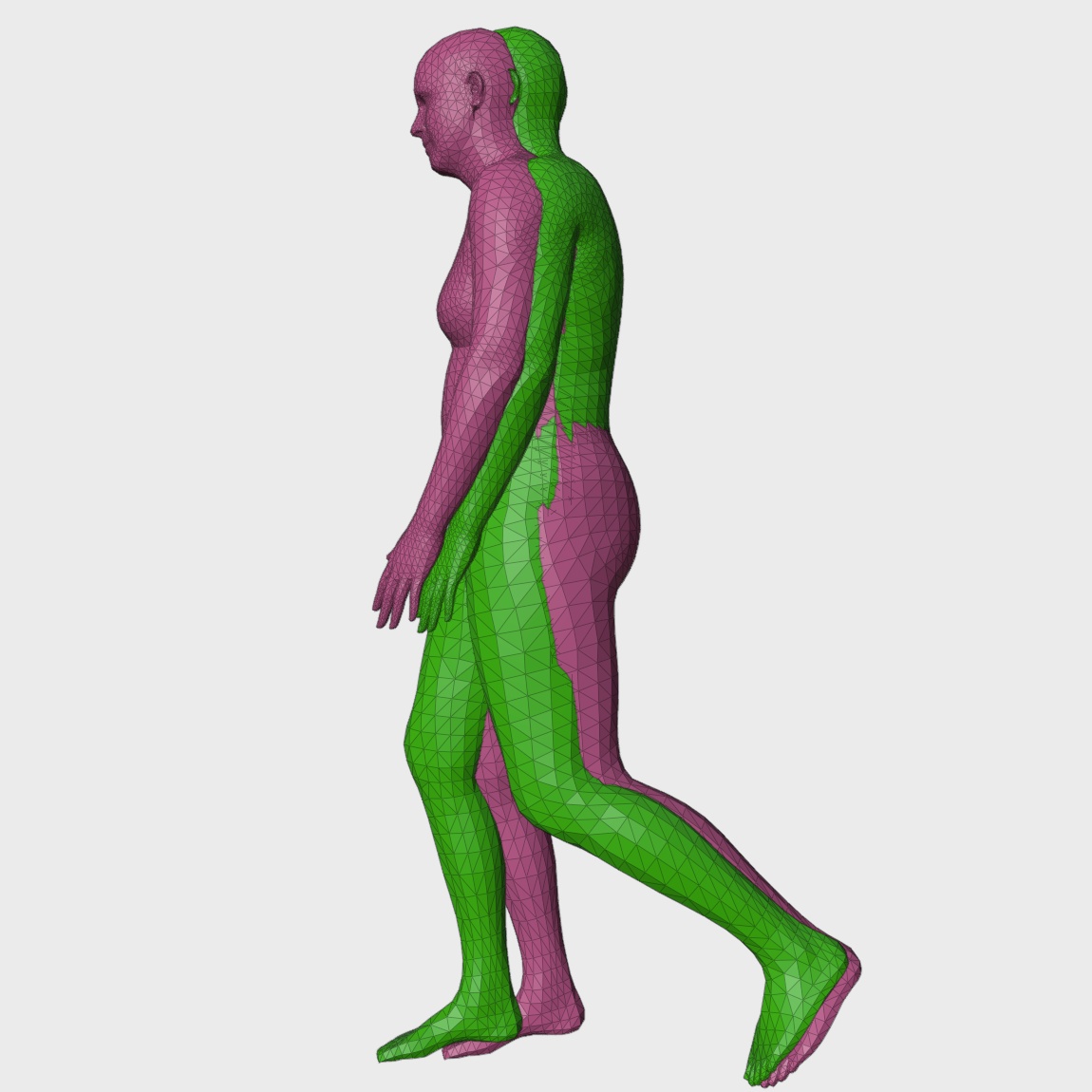}
\includegraphics[width=0.104\textwidth]{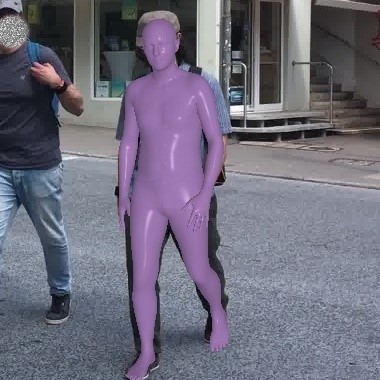}
\includegraphics[width=0.104\textwidth]{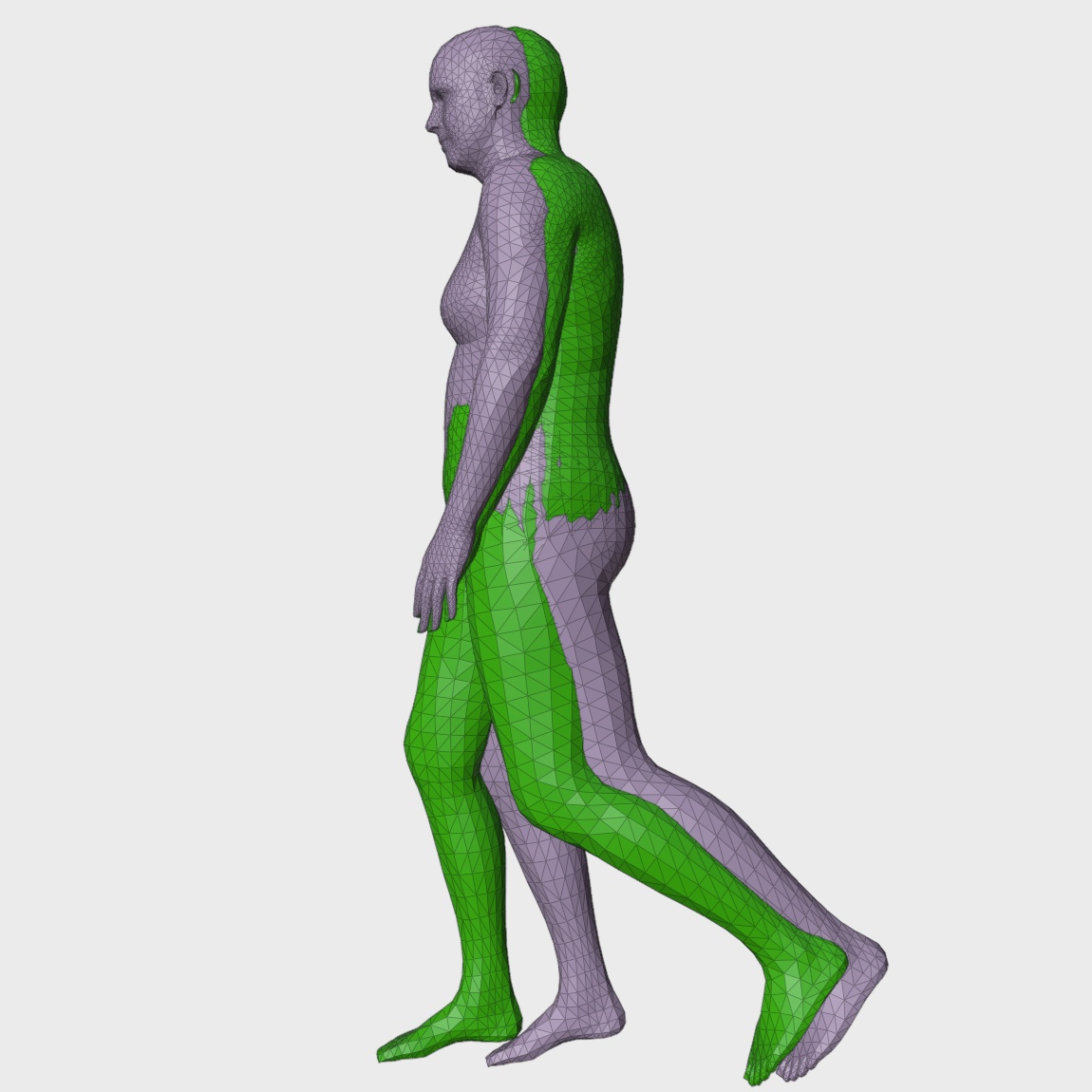}
\includegraphics[width=0.104\textwidth]{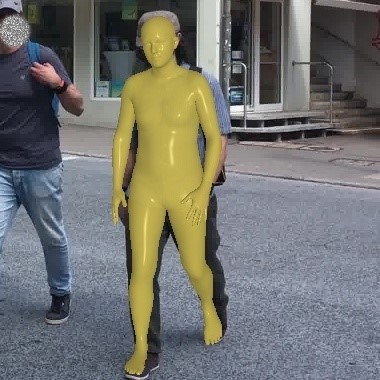}
\includegraphics[width=0.104\textwidth]{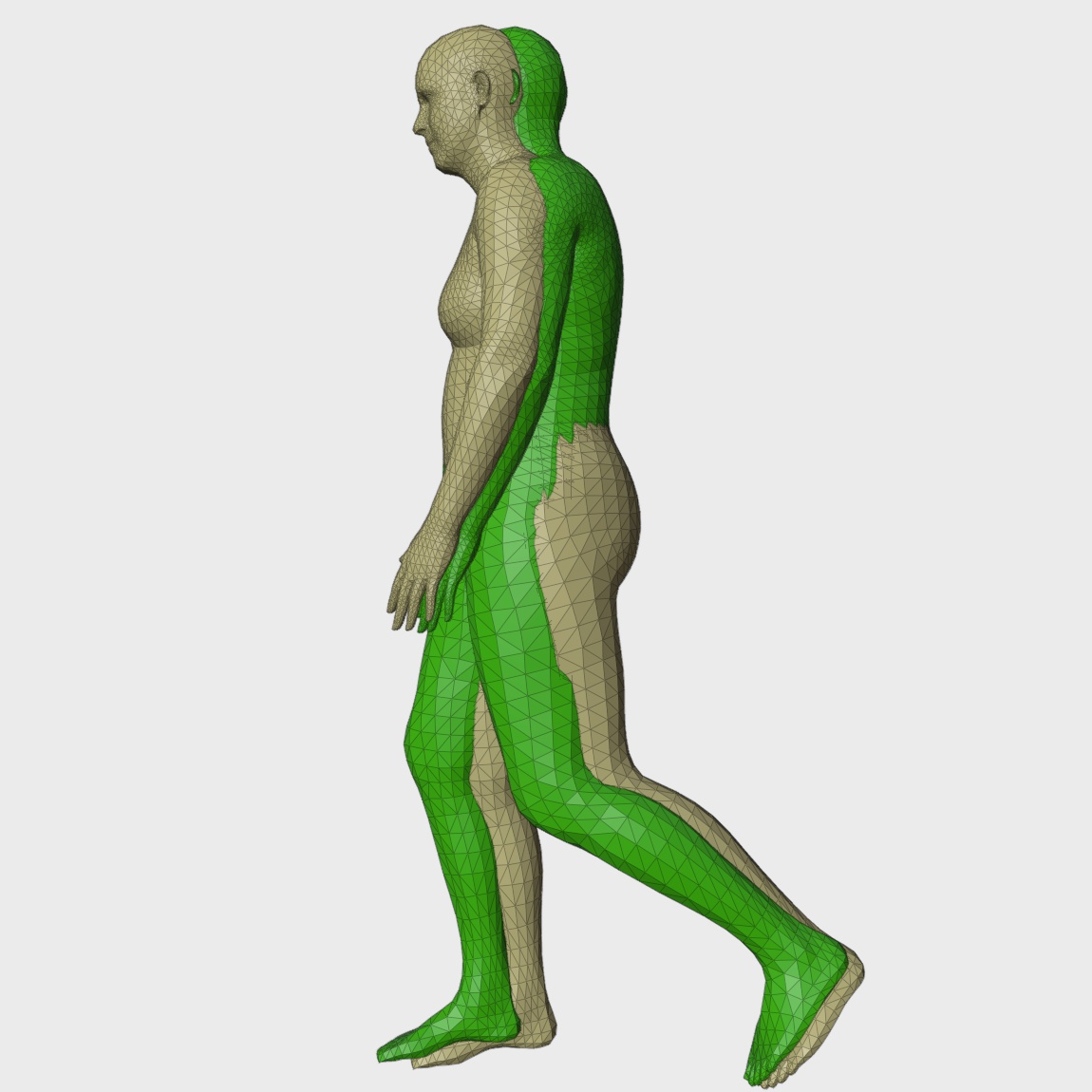} \\
\includegraphics[width=0.104\textwidth]{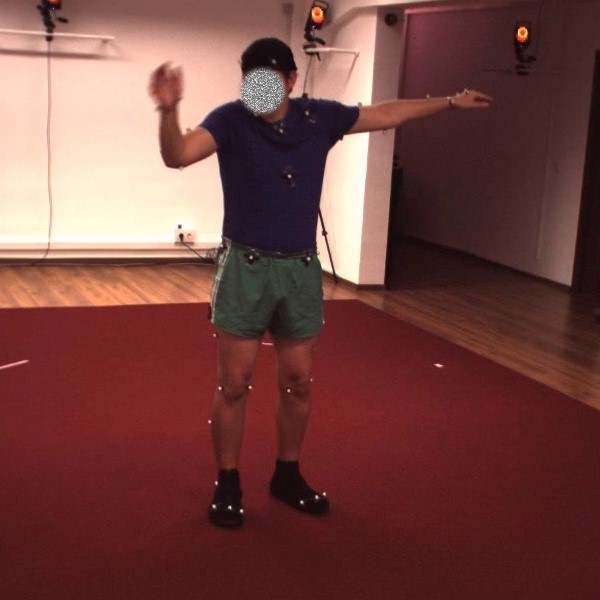}
\includegraphics[width=0.104\textwidth]{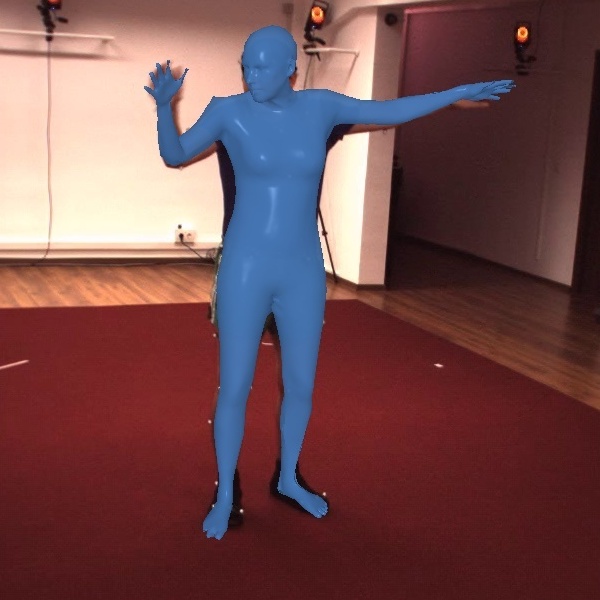}
\includegraphics[width=0.104\textwidth]{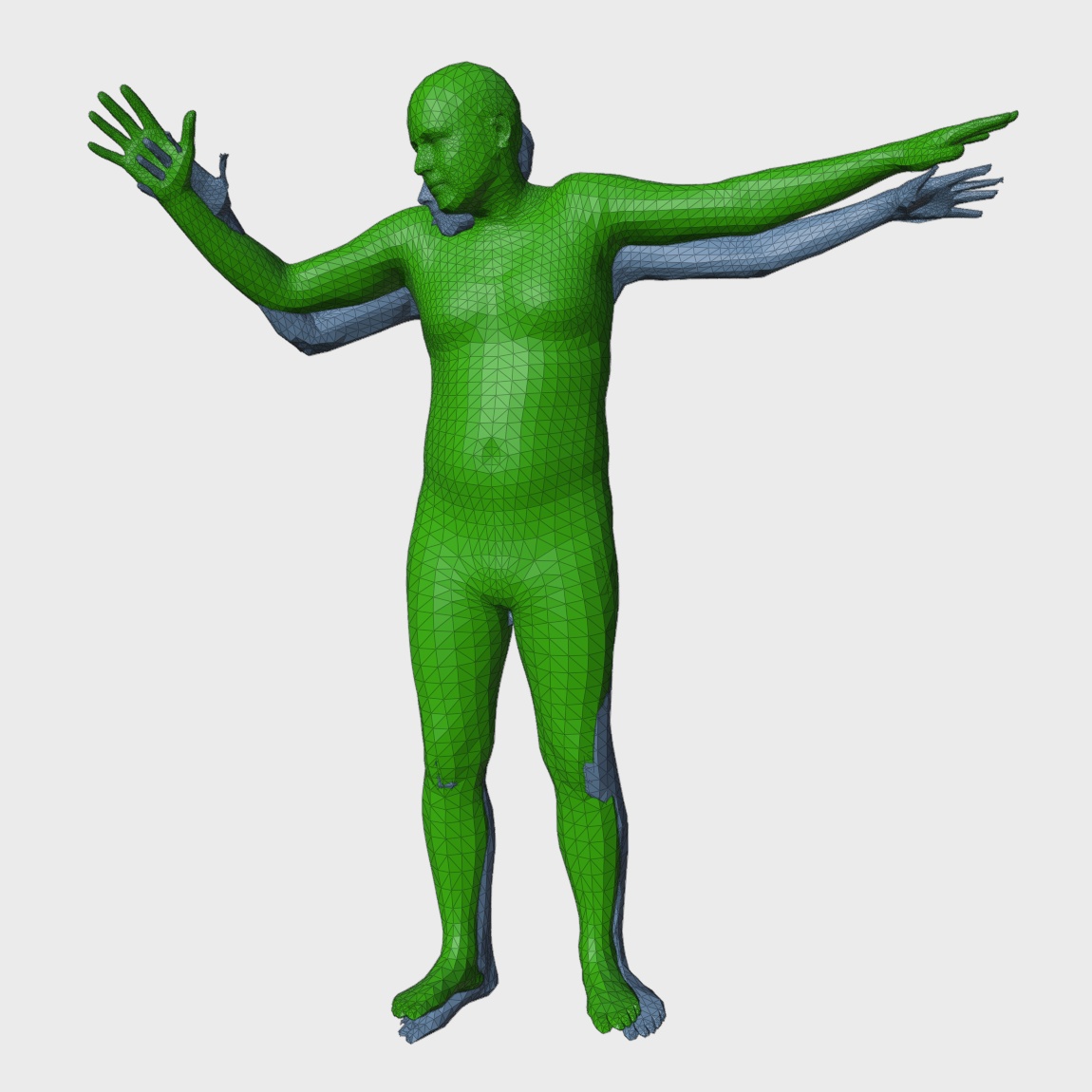}
\includegraphics[width=0.104\textwidth]{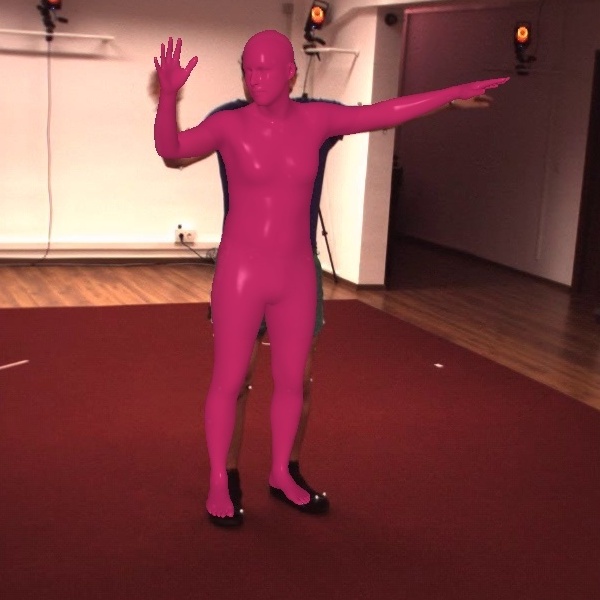}
\includegraphics[width=0.104\textwidth]{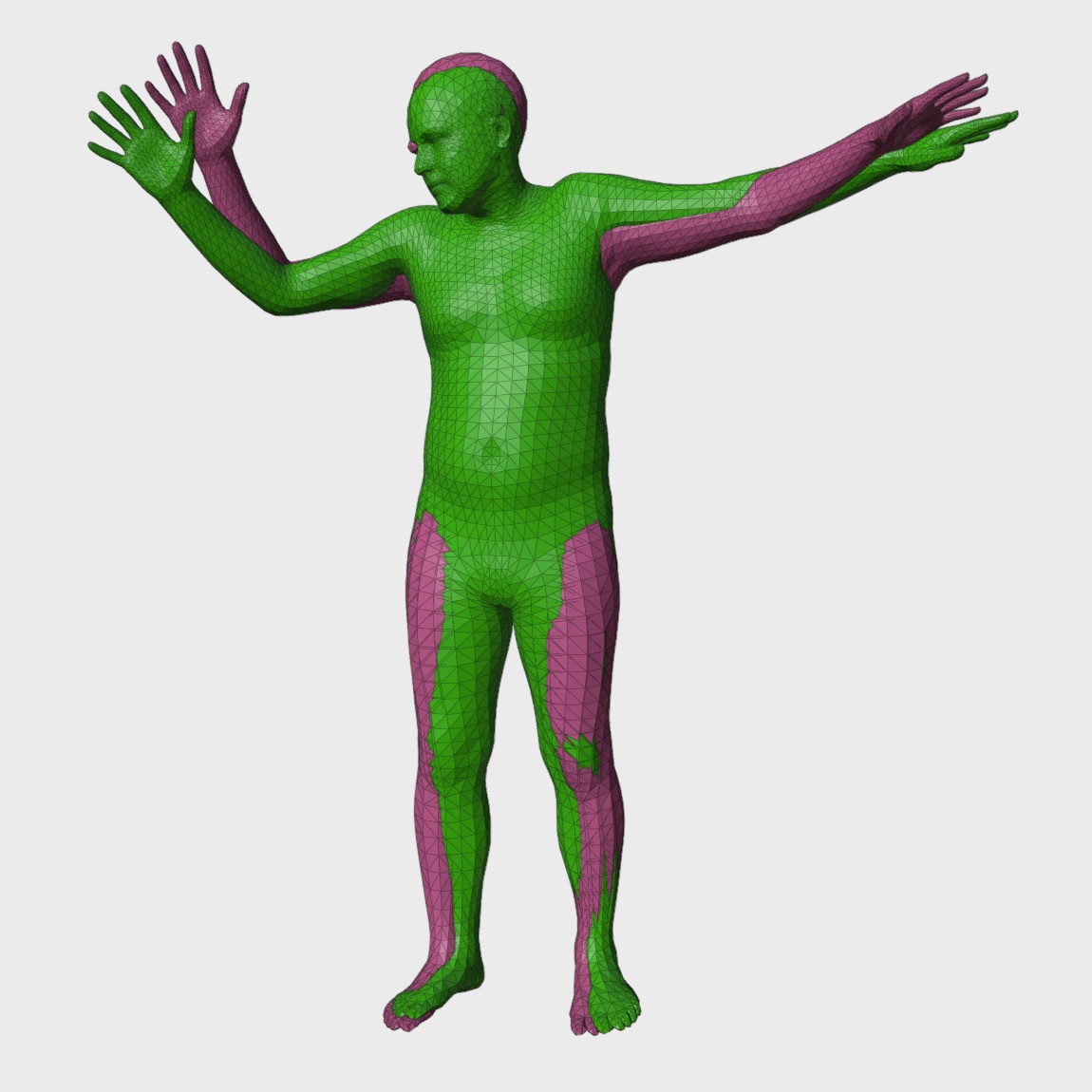}
\includegraphics[width=0.104\textwidth]{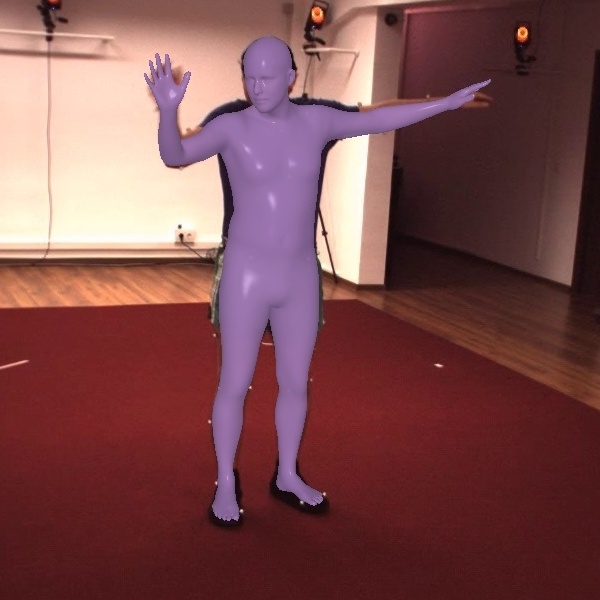}
\includegraphics[width=0.104\textwidth]{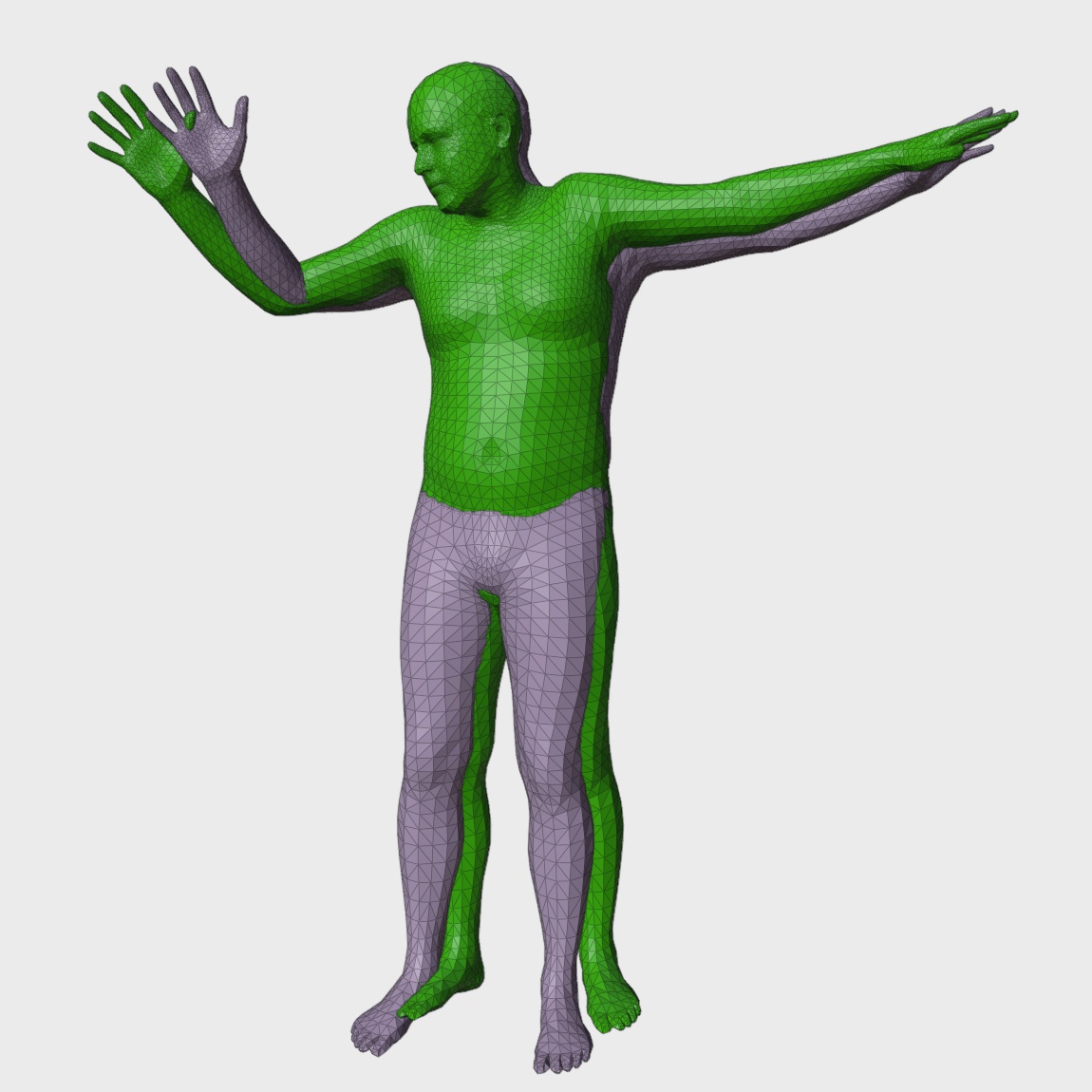}
\includegraphics[width=0.104\textwidth]{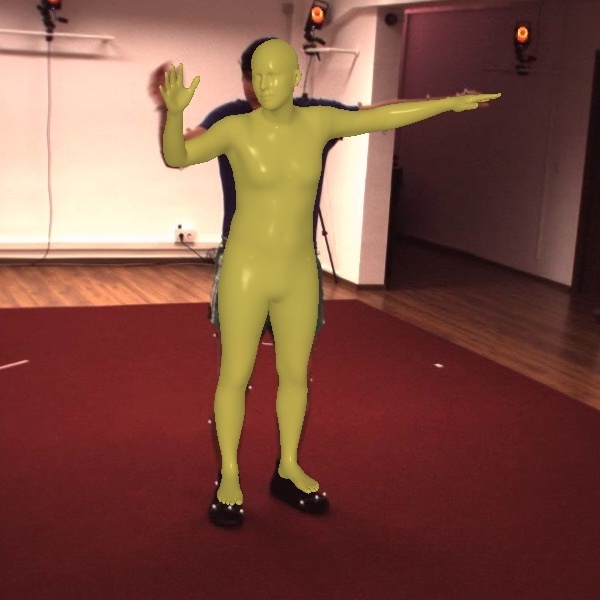}
\includegraphics[width=0.104\textwidth]{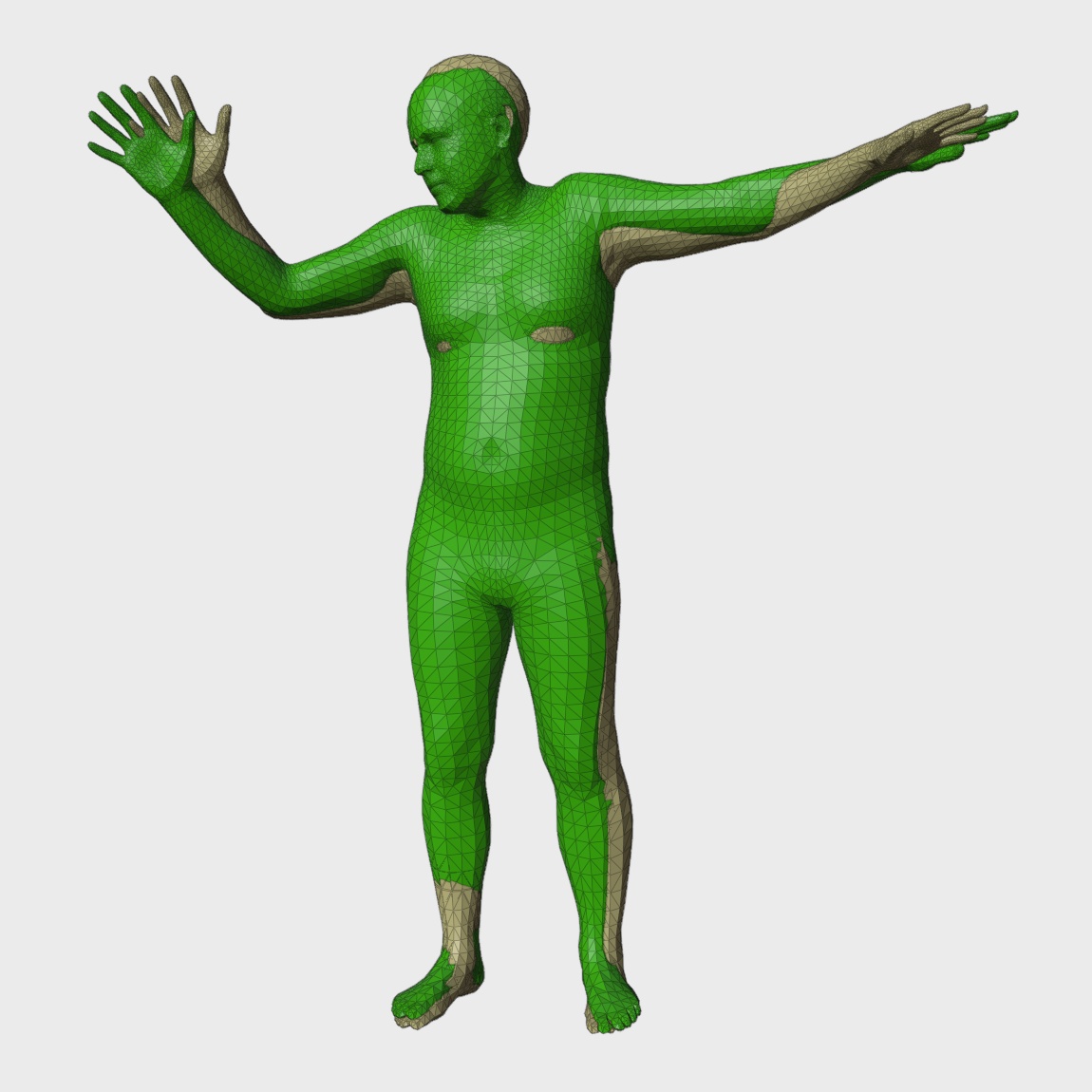} \\
\includegraphics[width=0.104\textwidth]{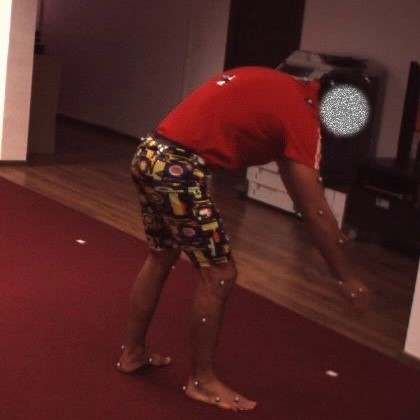}
\includegraphics[width=0.104\textwidth]{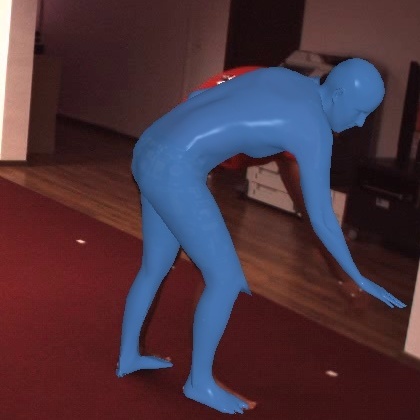}
\includegraphics[width=0.104\textwidth]{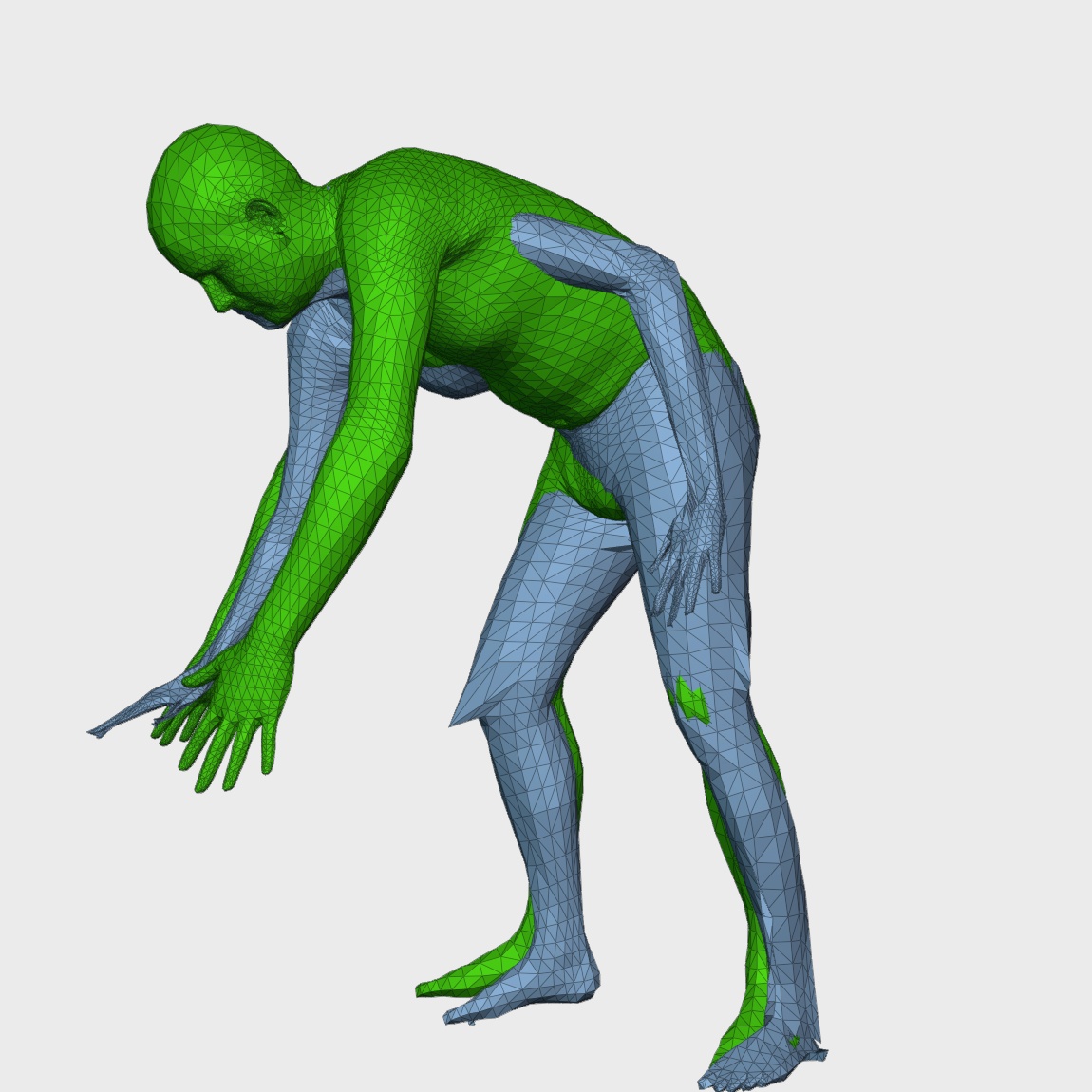}
\includegraphics[width=0.104\textwidth]{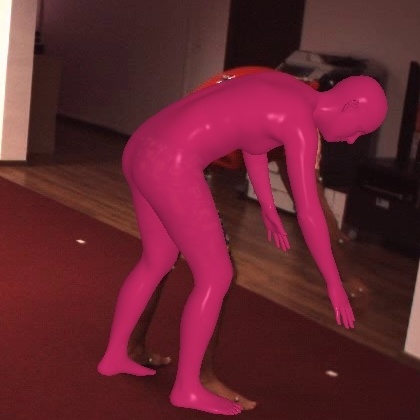}
\includegraphics[width=0.104\textwidth]{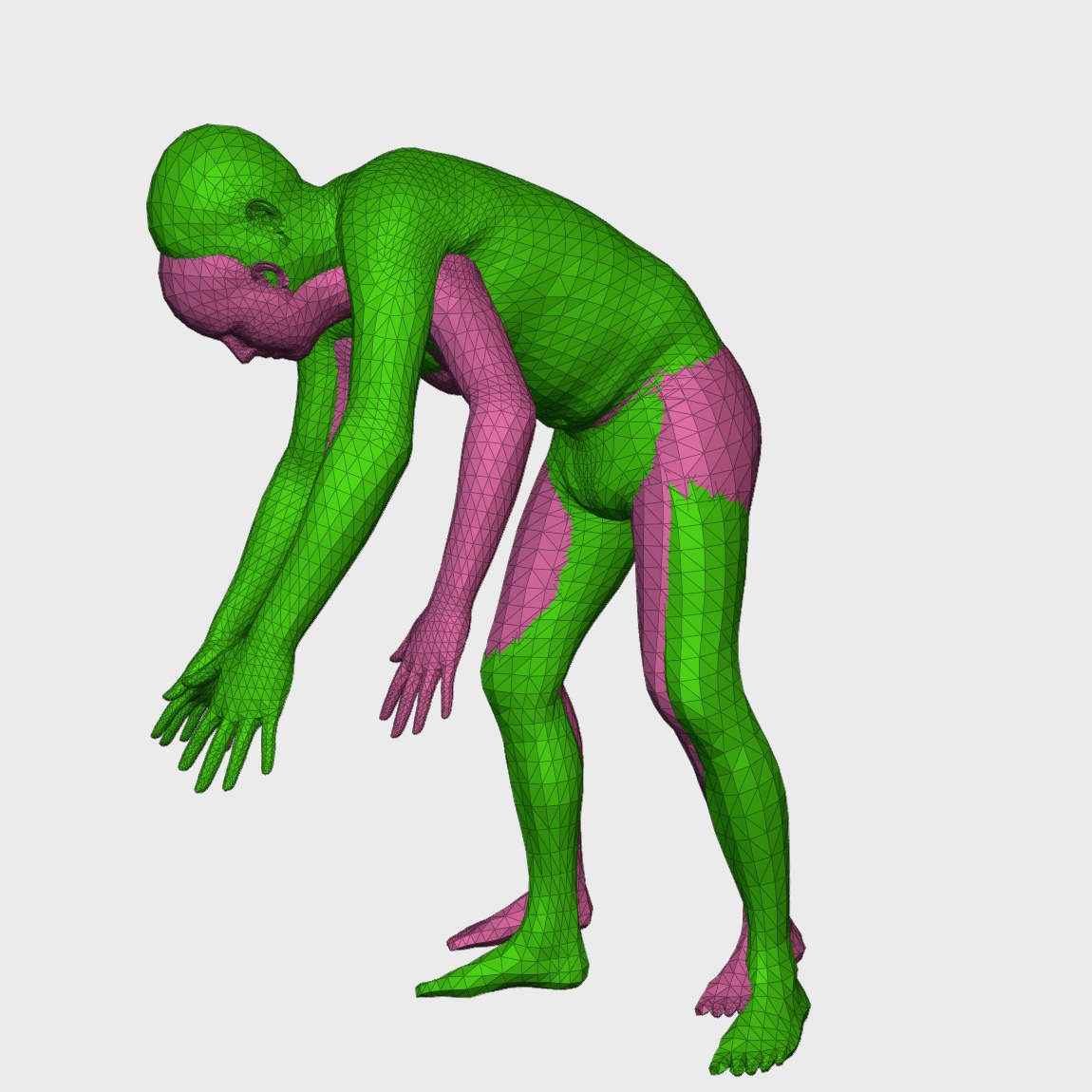}
\includegraphics[width=0.104\textwidth]{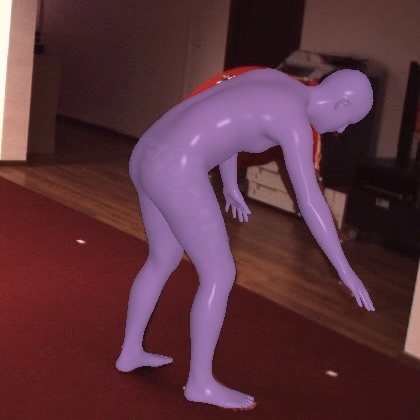}
\includegraphics[width=0.104\textwidth]{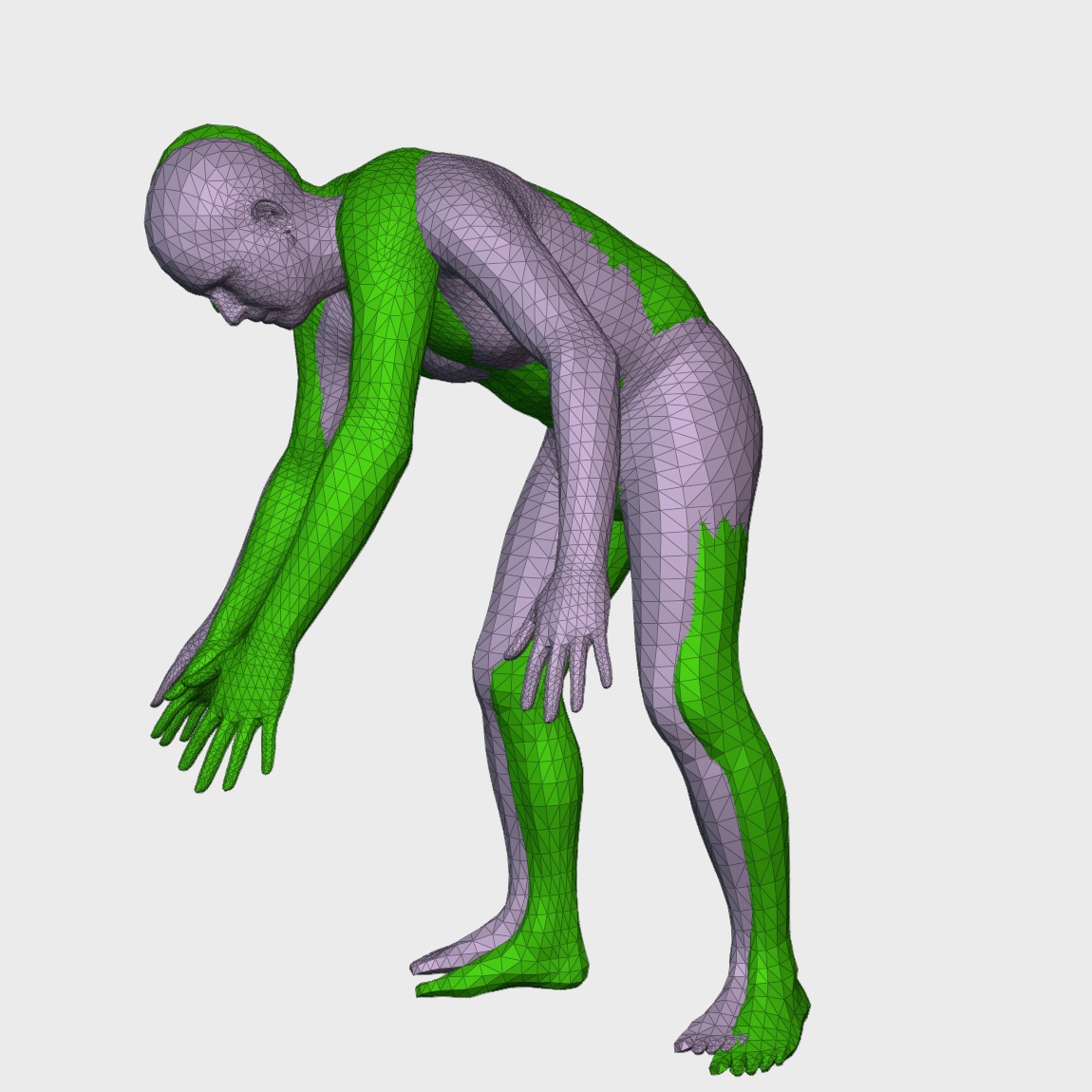}
\includegraphics[width=0.104\textwidth]{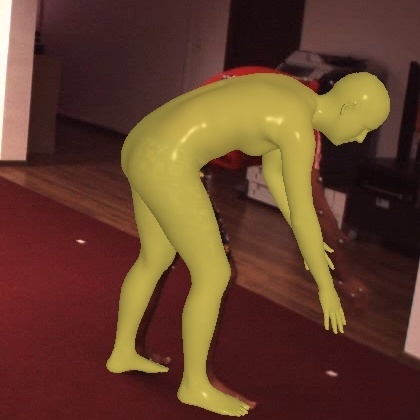}
\includegraphics[width=0.104\textwidth]{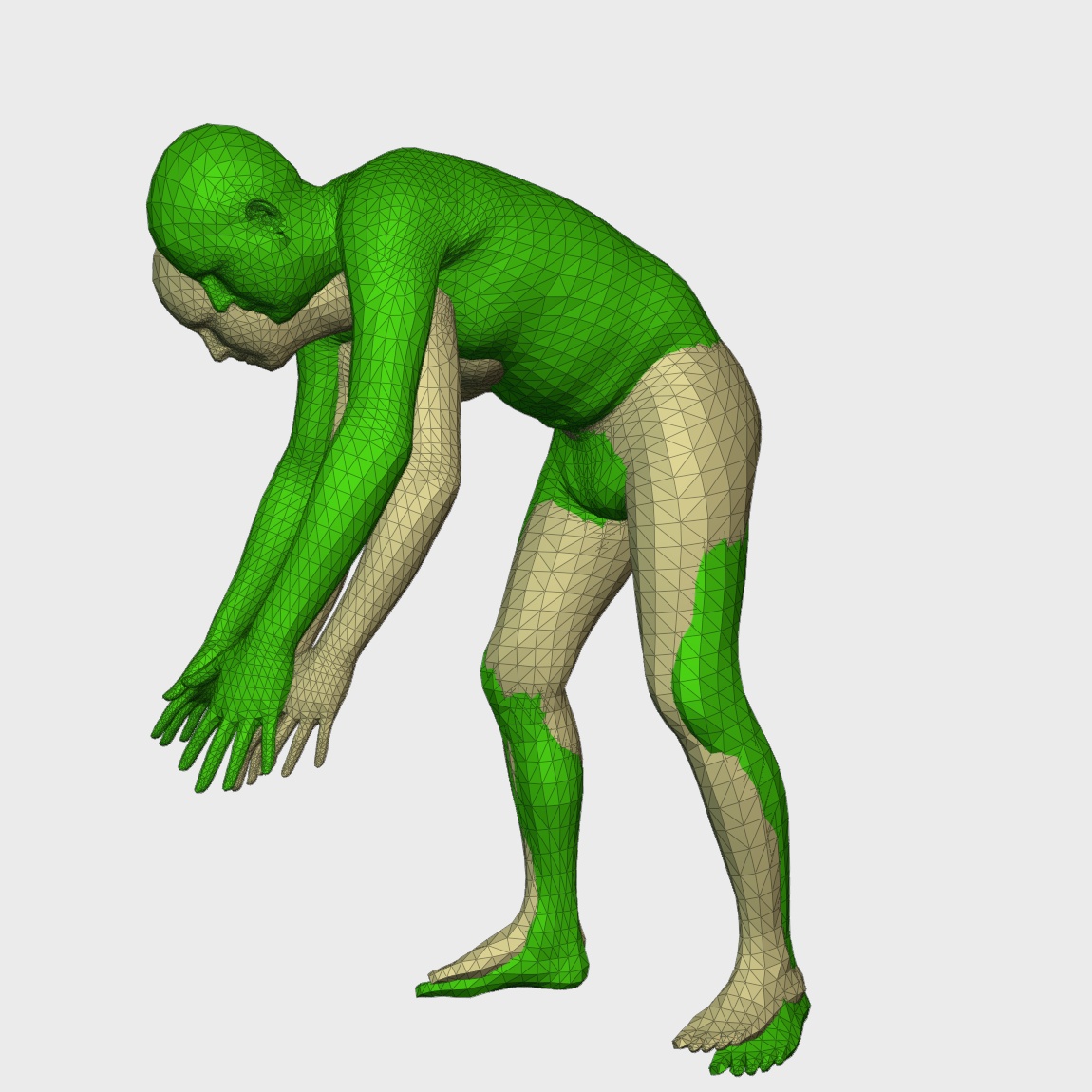}\\
\flushleft \small \quad Input ~\quad FastMETRO \cite{cho2022cross} \qquad \quad  CLIFF \cite{li2022cliff} \qquad \quad ReFit \cite{wang2023refit} \qquad \qquad Ours \\
\caption{\textbf{Qualitative comparison with SOTA approaches.} Our results align better with the GT mesh (green) than other results.}
\label{fig:big-figure-example1}
\end{figure*}

\subsection{Comparison to Prior Arts}
\label{subsec:Comparison}
Table \ref{tab:main_results_} provides quantitative comparisons with SOTA approaches. 
We compare with IK-based approaches \cite{li2021hybrik,li2023niki}, iterative fitting approaches \cite{zhang2021pymaf,zhang2023pymaf,wang2023refit}, Transformer-based approaches \cite{cho2022cross,yoshiyasu2023deformable}, and approaches improving camera \cite{li2022cliff,wang2023zolly}, etc.  
As seen, our method, either with a HRNet backbone or with a ResNet backbone, has better performance on the two evaluation datasets than the corresponding compared approaches. Please pay attention to the comparison between our method and CLIFF \cite{li2022cliff}, as our method is implemented based on CLIFF. Taking the backbone of HRNet-W48 as an example, the margin on MPJPE between CLIFF and ours is 5mm, which is a large improvement considering CLIFF is a very strong baseline. When compared with Zolly \cite{wang2023zolly} and NIKI~\cite{li2023niki}, our method works well in terms of all the three evaluation metrics. Zolly and NIKI are competitive in terms of PA-MPJPE but not MPJPE or PVE. Our method performs well on both testing datasets, while approaches such as PLIKS~\cite{shetty2023pliks} and ReFit~\cite{wang2023refit} show advantages on 3DPW but not Human3.6M. 

\begin{figure}[t]\scriptsize
	\begin{minipage}[t]{0.40\linewidth}\centering
			\includegraphics[width=0.45\columnwidth]{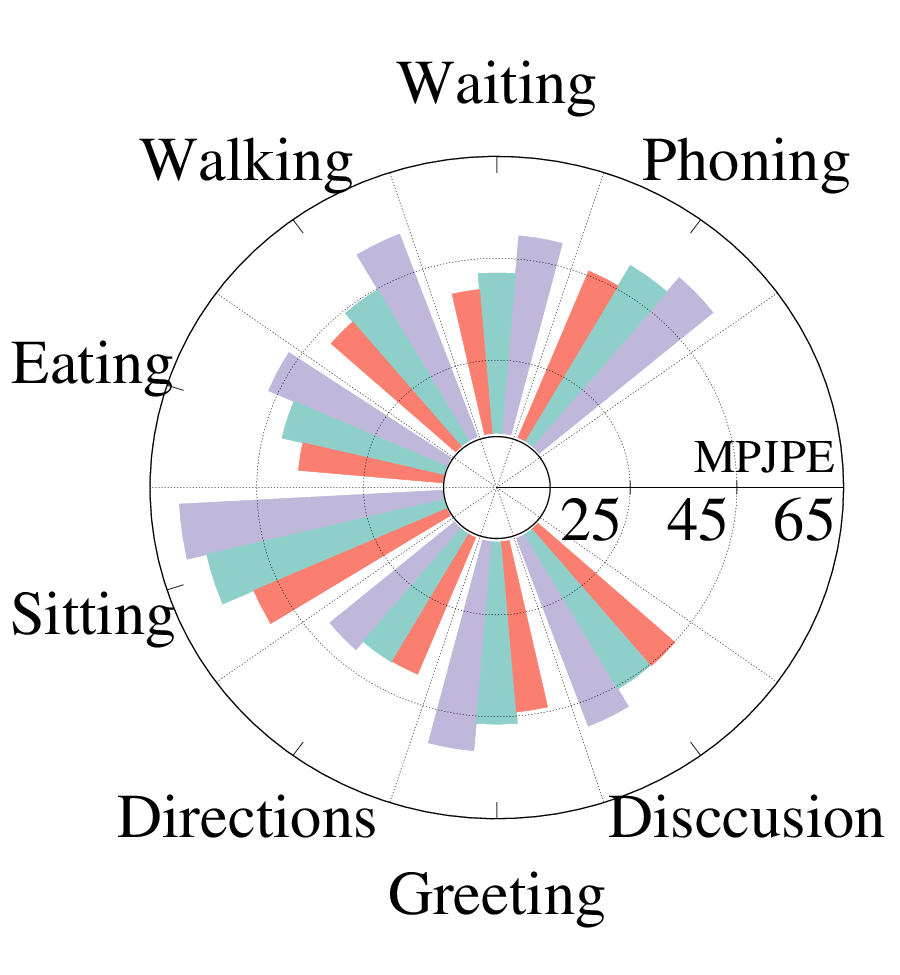}
			\includegraphics[width=0.45\columnwidth]{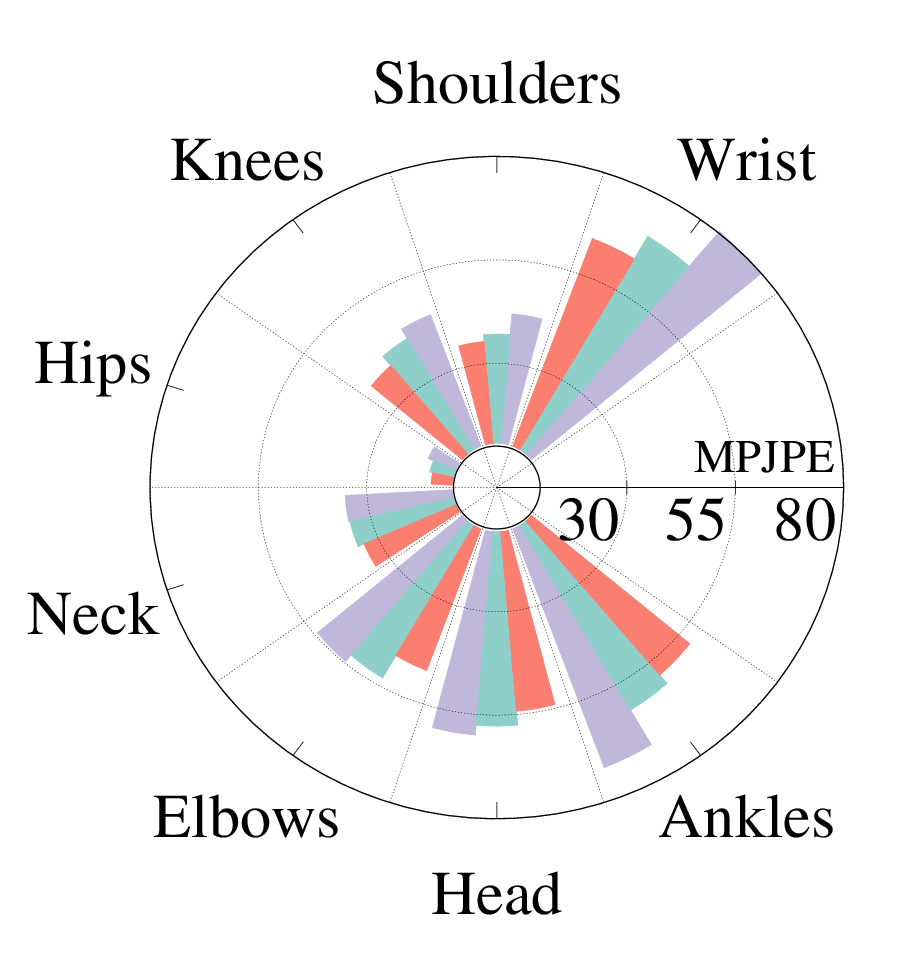}
			\\
			\includegraphics[width=0.92\columnwidth]{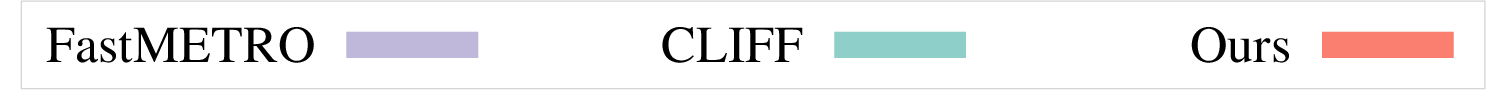}
		\caption{\textbf{Per action (left) or joint (right) MPJPE comparison} with FastMETRO \cite{cho2022cross} and CLIFF \cite{li2022cliff} on Human3.6M.}
		\label{fig:per-action-joint}
	\end{minipage}
	\hfill
	\begin{minipage}[t]{0.55\linewidth}\centering
			\centering
			\includegraphics[width=0.315\columnwidth]{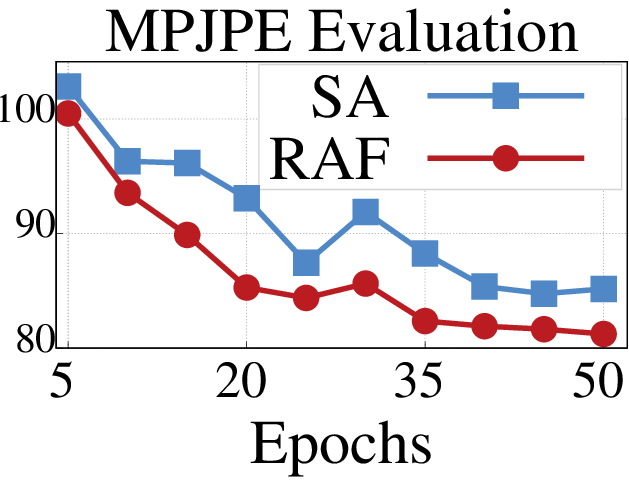}
			\includegraphics[width=0.315\columnwidth]{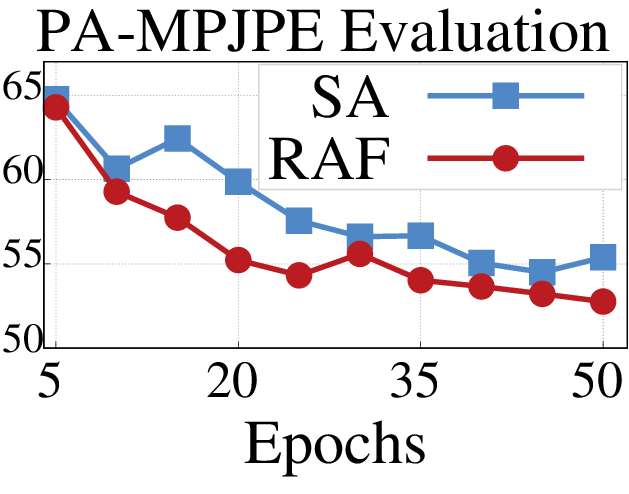}
			\includegraphics[width=0.315\columnwidth]{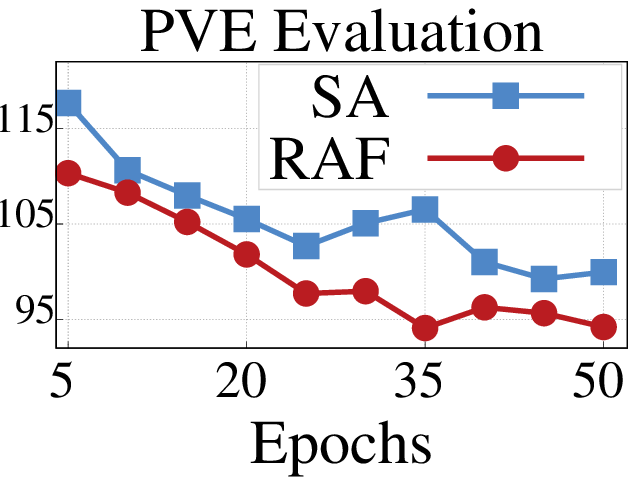}	
		\caption{\textbf{Comparison between self attention (SA) fusion and our relative-relation-based fusion (RAF)}. Accuracy at different training epochs are shown.}
		\label{fig:comparison between sa and RAF}
	\end{minipage}
\end{figure}

Figure~\ref{fig:big-figure-example1} shows qualitative comparisons between our method and SOTA approaches. The shown cases are challenging, containing either complex poses or showing occlusions by other body parts. For these cases, our estimated meshes resemble the GT (green color) better than results of the compared approaches. Figure~\ref{fig:per-action-joint} shows the per action and per joint comparisons on Human3.6M. Our method outperforms FastMETRO and CLIFF on all kinds of actions and joints.

\subsection{Ablation Study}
\label{subsec:Ablation}

In this section, we conduct ablation studies on the core design ideas of our method.  
All the following ablation studies are performed on the COCO training dataset and tested on 3DPW following previous literature \cite{kocabas2021pare, li2022cliff}, if not otherwise specified.

\textbf{Ablation on Design Components.}
Our method is composed of three major components: RoI-aware fusion module (RAF), camera consistency loss ($\mathcal{L}_{cam}$) and contrastive loss ($\mathcal{L}_{cont}$). To show the effect of each component, we remove each of them at a time while maintaining the other two components (ablations of removing two components are provided in the Supp.). Table~\ref{tab:ablation-components} shows that removing any component incurs an apparent performance drop. Especially, when we drop $\mathcal{L}_{cam}$, the MPJPE increases by 3mm while PA-MPJPE stays low, indicating that $\mathcal{L}_{cam}$ assists to predict more accurate mesh orientation by improving cameras. 

\textbf{Importance of Relative Relation and Positional Encoding.} As discussed in Section~\ref{subsec: RAF}, we rely on relative relation for the RoI-aware feature fusion (also denoted as RAF), and the relation is computed based on the positional encoding (PE) of the bounding boxes. Both of them are critical to our method, as shown in Table~\ref{tab:importance-of-rpe}: 
(1) $\mathbf{h}_*\oplus$ NULL: concatenating nothing, \ie, using only feature $\mathbf{h}_*$ for computing relation weights, where $*$ is a number in $[1,M]$. (2) $\mathbf{h}_*\oplus\gamma(\mathbf{B}_*)$: simply concatenating PE of the corresponding boundingbox. (3) $\mathbf{h}_*\oplus\gamma_{m*}$: concatenating relative PE $\gamma_{m*}$ for computing $m^{th}$ fused feature. We also test the above three setups with different length of PE, denoted as $L$.
As seen, using relative PE with $L=32$ yields the best results. 

We also implemented RAF by performing self attention \cite{vaswani2017attention} on $M$ tokens of $\{\mathbf{h}_m\oplus\gamma(\mathbf{B}_m)\}_{m=1}^M$. Here we can only concatenate PE but not relative PE, since there is only $M$ tokens but we have $M^2$ relative relations (see supplemental material for details). Results are shown in Figure~\ref{fig:comparison between sa and RAF}, where our relative-PE based scheme \ie, RAF, outperforms the self-attention approach.

\textbf{Number of RoIs.} 
We conduct an ablation study that gradually increases the number of input RoIs in Table \ref{tab:ablation-crop-number}. 
As seen, the accuracy is consistently increased as the number of input RoIs increases. Experiments of inputting 6 or more RoIs are not conducted due to memory limit. 
We find that as the RoI number increases, the loss $\mathcal{L}_{2D}$ in Eq.~\ref{eq:loss_overall} increases while higher regression accuracy can be obtained. 
This indicates that inputting more RoIs may prevent the network from over-fitting. 

\textbf{Inferring Speed.} We report the inferring speed in Table \ref{tab:ablation-crop-number}. As the number of RoIs increases, the inferring speed is just slightly decreased. With 5 RoIs, our method processes 55.6 frames per second, which is fast.

\begin{table}[t]\scriptsize
	\begin{minipage}[t]{0.47\linewidth}\centering
    \caption{\textbf{Ablation on core components of our method.}}
		\begin{tabular*}{\textwidth}{p{0.8cm}<{\centering}p{0.8cm}<{\centering}p{1.1cm}<{\centering}|p{1.0cm}<{\centering}p{1.5cm}<{\centering}}
			\toprule
			  RAF & $\mathcal{L}_{cam}$ & $\mathcal{L}_{cont}$ & MPJPE & PA-MPJPE \\
			\midrule
			\ding{55}  & \ding{55}  & \ding{55} & 87.0 & 55.8\\
			\ding{55} &\checkmark & \checkmark & 83.8 & 54.8\\
			\checkmark & \ding{55} & \checkmark & 83.1 & 52.0\\
            \checkmark &\checkmark & \ding{55} & 83.4 & 53.3\\
			\rowcolor{mygray}
			\checkmark &\checkmark & \checkmark & \textbf{80.8} & \textbf{51.9}
			\\
			\bottomrule
		\end{tabular*}
		\label{tab:ablation-components}
	\end{minipage}
    \begin{minipage}[t]{0.51\linewidth}\centering
    \caption{\textbf{Importance of Relative Relation and Positional Encoding (PE).}}
			\begin{tabular*}{\columnwidth}{p{1.7cm}<{\raggedright}p{1.55cm}<{\centering}|p{1.05cm}<{\centering}p{1.5cm}<{\centering}}
				\toprule
				Concatenation  &  $L$ of PE  & MPJPE           & PA-MPJPE      \\
				\midrule
				$\mathbf{h}_*~\oplus$ NULL            &$-$                        & 98.7          & 56.9          \\
				$\mathbf{h}_*\oplus\gamma(\mathbf{B}_*)$ & $L = $~32                 & 87.6          & 55.0          \\
				$\mathbf{h}_*\oplus\gamma_{m*}$ & $L =~$~0                 & 81.6          & 52.9          \\
				\rowcolor{mygray}
				$\mathbf{h}_*\oplus\gamma_{m*}$ & $L = $~32                 & \textbf{80.8}   & \textbf{51.9} \\
				$\mathbf{h}_*\oplus\gamma_{m*}$ & $L = $~64                 & 81.1          & 52.3         \\
				\bottomrule
			\end{tabular*}
			\label{tab:importance-of-rpe}
	\end{minipage}
\end{table}

\begin{table}[!t]\scriptsize
\centering
\caption{\textbf{Ablation on number of input RoIs.} Five is the default.}
\begin{tabular*}{0.7\columnwidth}{p{2.89cm}<{\raggedright}|p{1cm}<{\centering}|p{1cm}<{\centering}|p{1cm}<{\centering}|p{1cm}<{\centering}|p{1cm}<{\centering}}
\hline
RoI Num  & 1 RoI    & 2 RoIs   & 3 RoIs   & 4 RoIs  & \cellcolor{mygray}5 RoIs            \\
\hline
MPJPE~(mm)   & 84.6 & 82.9 & 81.2 & 82.2 & \cellcolor{mygray}\textbf{80.8} \\
PA-MPJPE~(mm) & 54.2 & 54.4 & 52.3 & 53.1 & \cellcolor{mygray}\textbf{51.9} \\
Final $\mathcal{L}_{2D} ~(\times 1e-3)$  & \textbf{1.6} & 1.8 & 1.9 & 2.3 &\cellcolor{mygray}2.4 \\
Inferring speed~$(fps)$ & \textbf{76.3} & 63.5 &61.7 & 57.3 &\cellcolor{mygray}55.6\\
\hline
\end{tabular*}
\label{tab:ablation-crop-number}
\end{table}

\subsection{Limitations} 
\begin{figure}[!t]
\centering
\includegraphics[width=0.192\columnwidth]{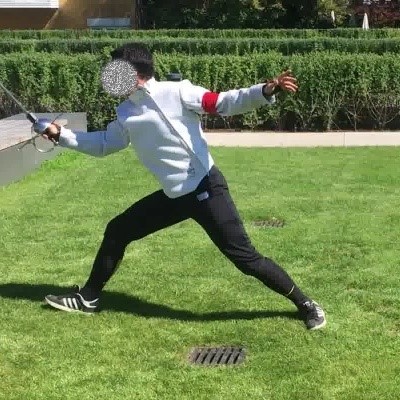}
\includegraphics[width=0.192\columnwidth]{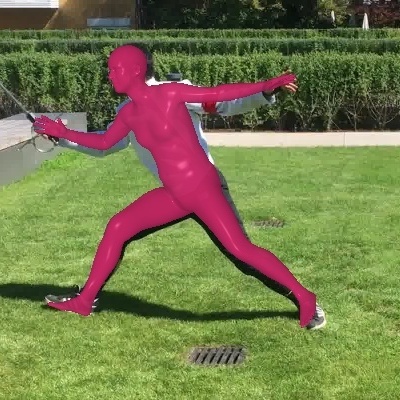}
\includegraphics[width=0.192\columnwidth]{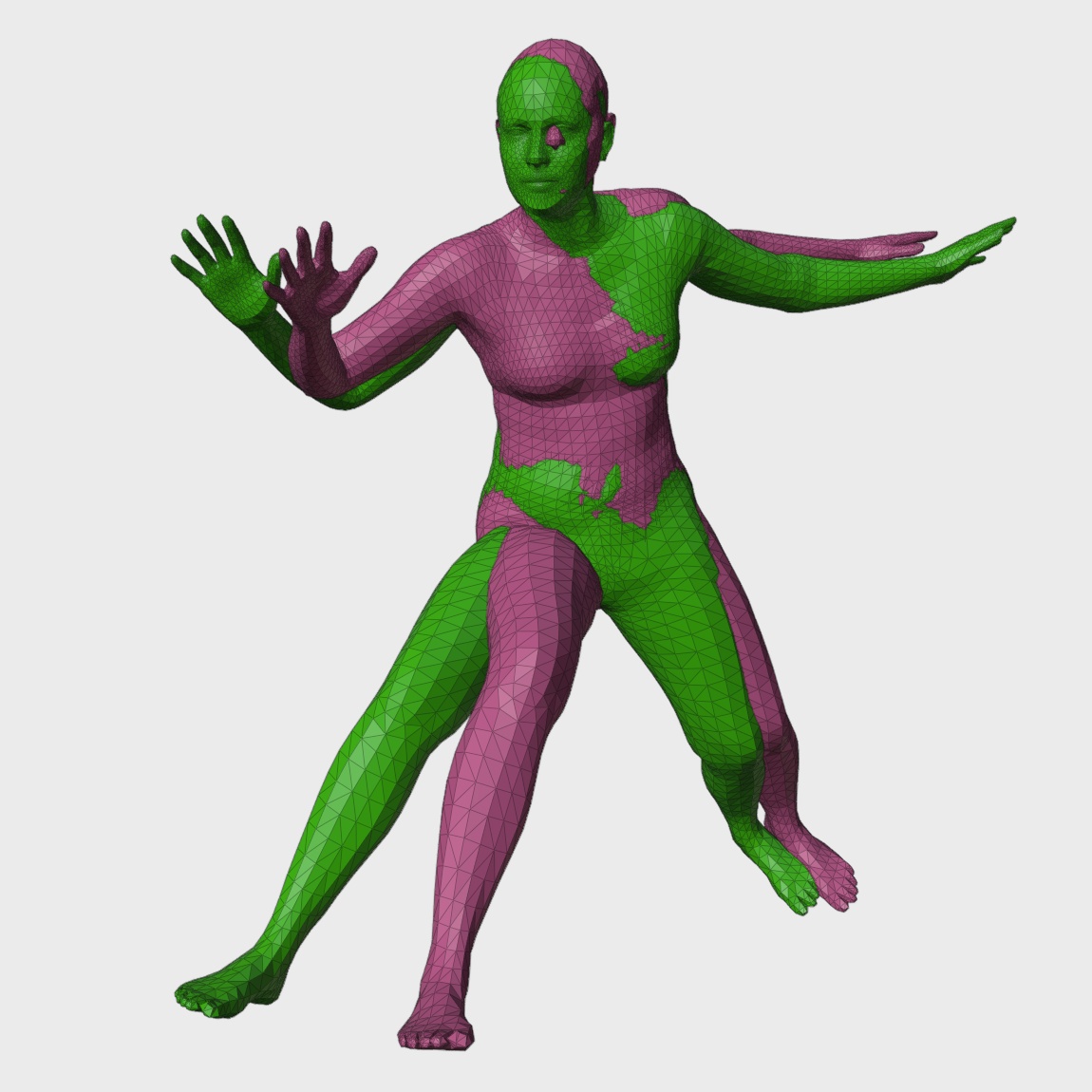}
\includegraphics[width=0.192\columnwidth]{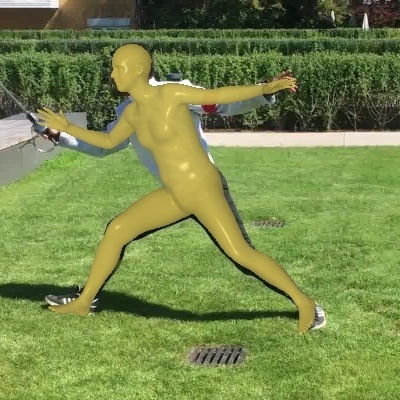}
\includegraphics[width=0.192\columnwidth]{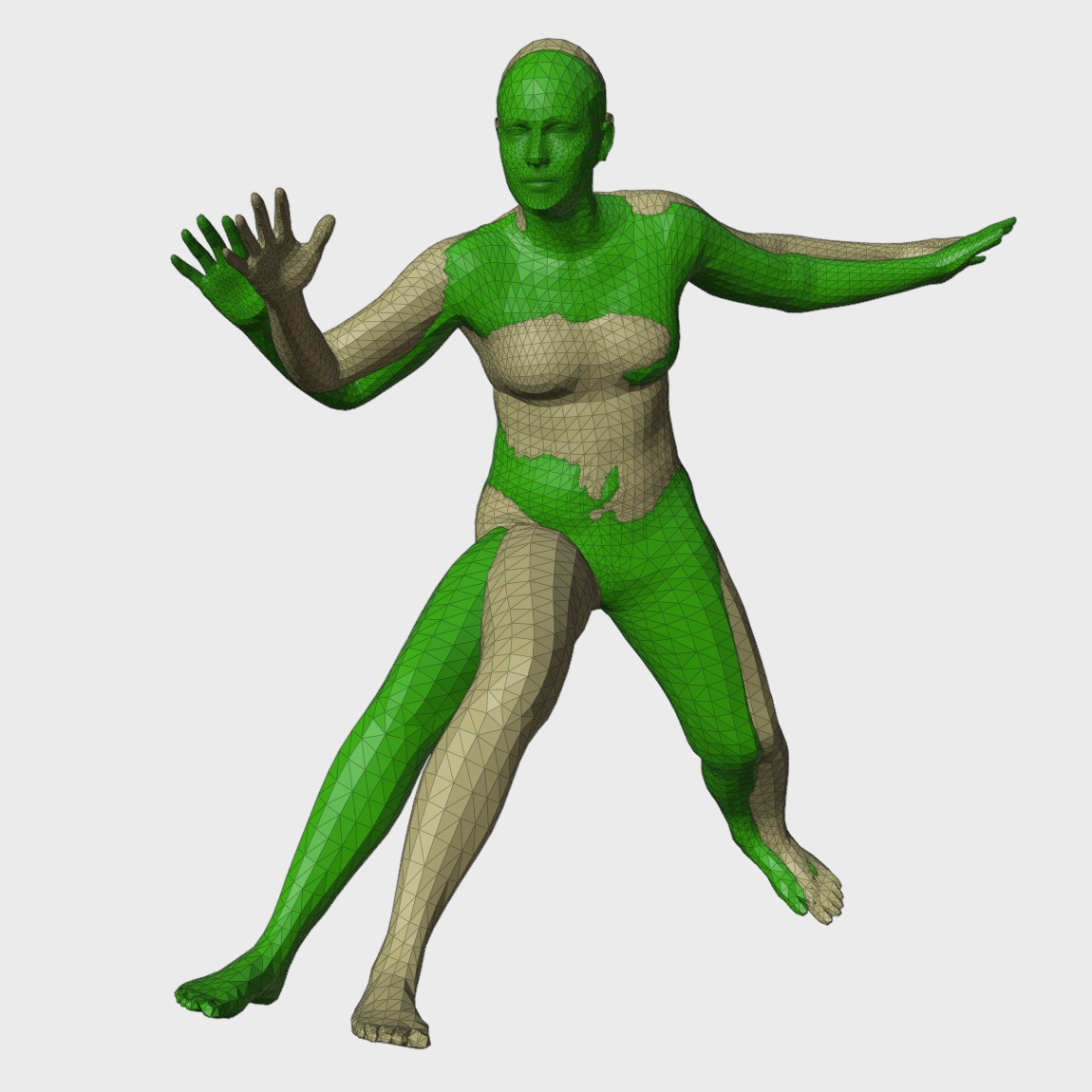} \\
\includegraphics[width=0.192\columnwidth]{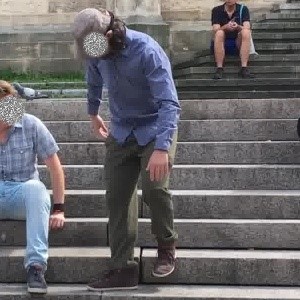}
\includegraphics[width=0.192\columnwidth]{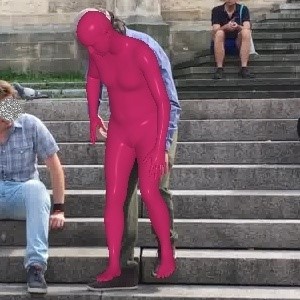}
\includegraphics[width=0.192\columnwidth]{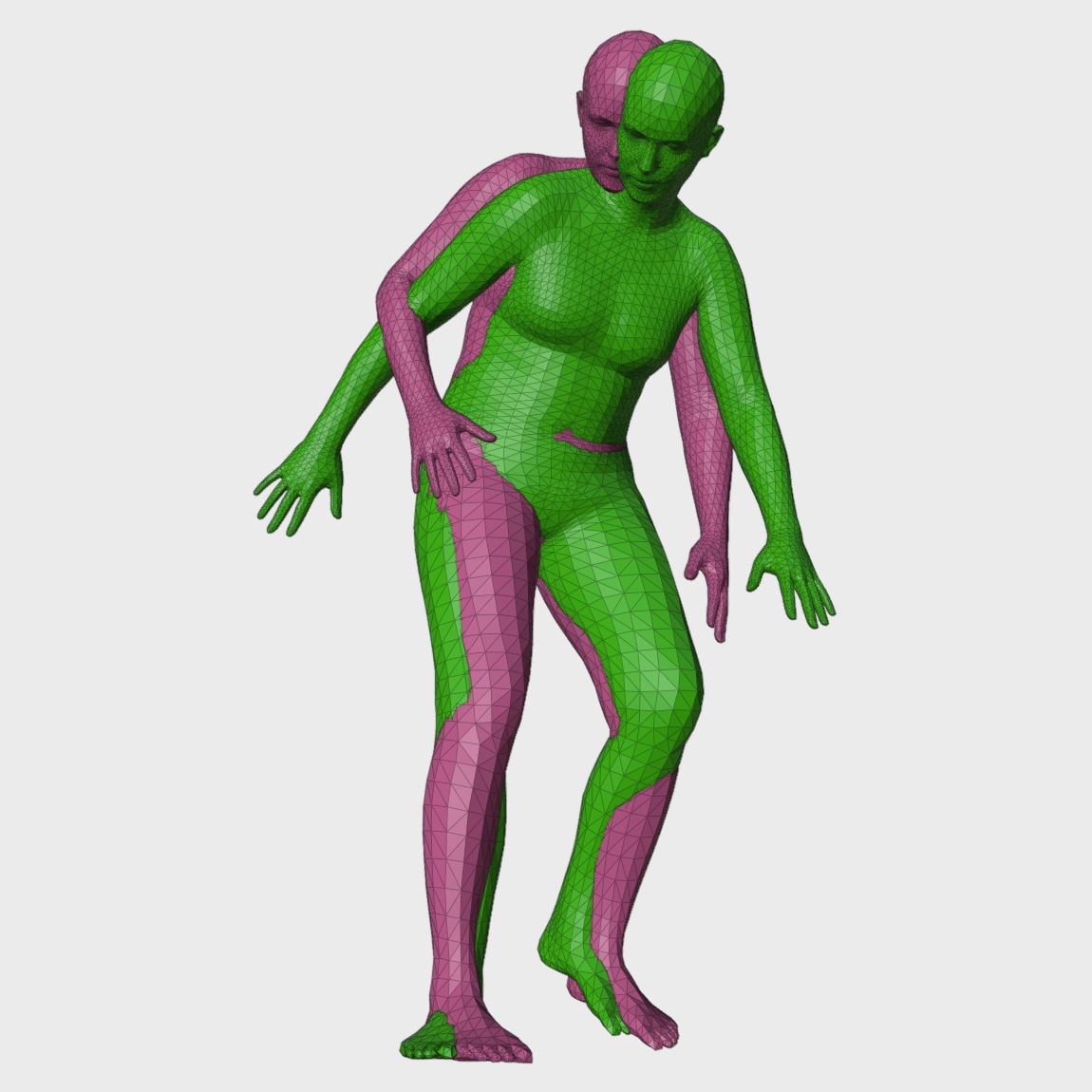}
\includegraphics[width=0.192\columnwidth]{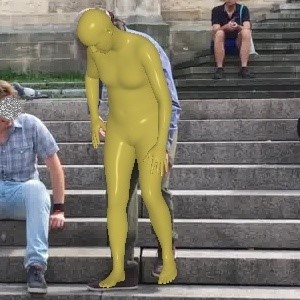}
\includegraphics[width=0.192\columnwidth]{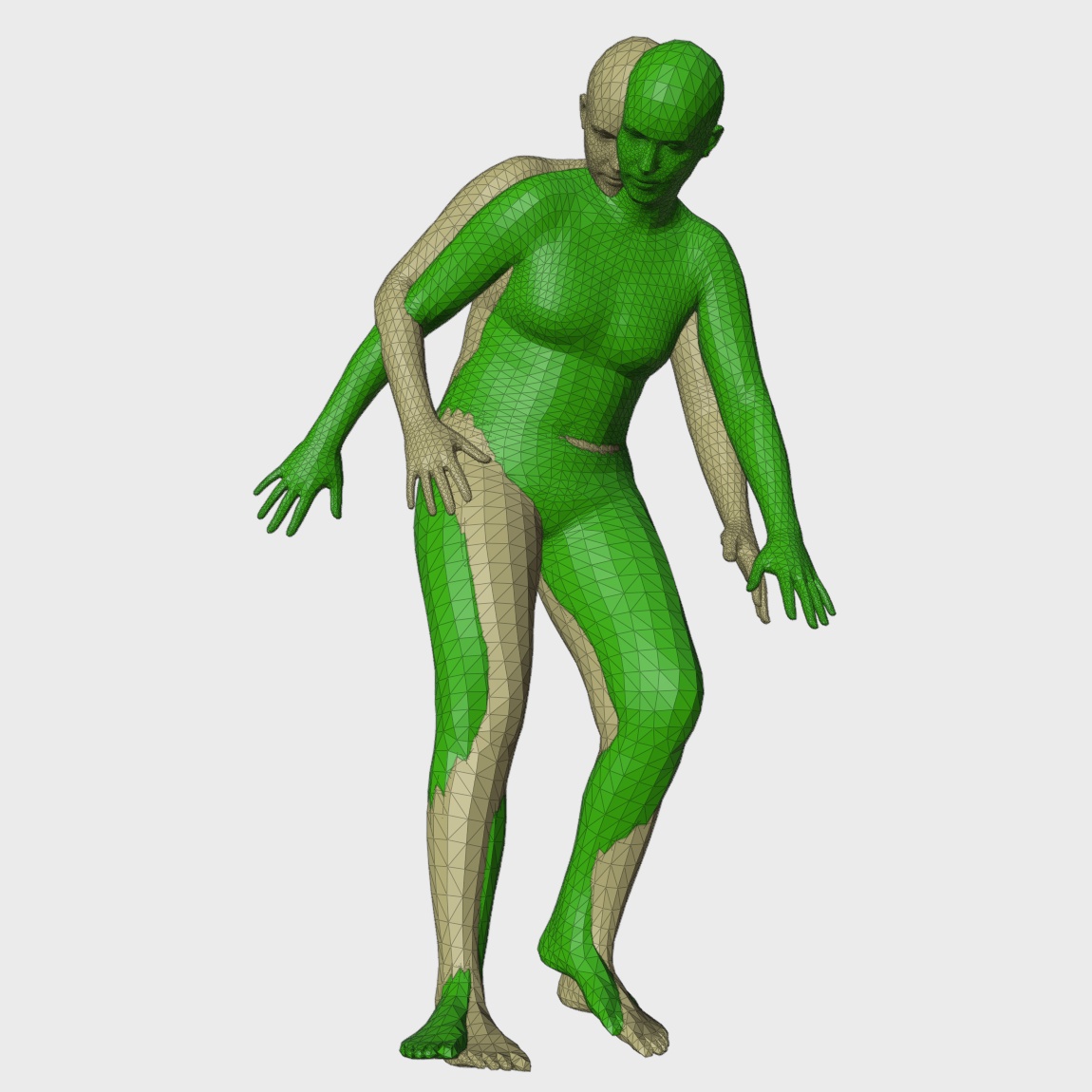} \\
\flushleft \small  \quad \qquad  Input \qquad \qquad \qquad \qquad  CLIFF \cite{li2022cliff} \qquad \qquad \qquad \qquad \qquad  Ours \\
\caption{\textbf{Failure examples.} Both our method and CLIFF produce nearly perfect results in the 2D image, but not in the 3D space.
}
\label{fig:failure-example}
\end{figure}

Figure \ref{fig:failure-example} shows failure cases, where both our method and CLIFF produce nearly perfect reprojection results without noticeable misalignment in the 2D image. But in the 3D space, some body parts of both methods deviate from the ground truth. These examples show that introducing multiple RoIs still cannot solve the ill-posed problem that there may be multiple 3D meshes matching with the same 2D configuration.

\section{Conclusion and Future Work}
\label{sec:summary}

This paper digs into the relation among different RoIs of the same person in an image for human mesh recovery. With the multiple RoIs indicated by different boundingboxes, we are able to design a multi-RoI fusion network to estimate reliable camera parameters, thanks to the additional visual information and pairwise relation provided by the multiple inputs. Specifically, we have exploited using relative-position-relation guided feature fusion, camera consistency loss and contrastive loss to take advantage of the information in multiple inputs as much as possible. We validate the effectiveness of each proposed component using experiments and prove our method has better regression accuracy than current SOTA approaches on popular benchmarks and datasets. In the future, it is valuable to investigate whether the proposed strategies are effective in multi-view or video-based HMR.

\section*{Acknowledgements}
This work was supported in part by the National Key Research and Development Program of China under grant 2022YFE0112200, in part by the Natural Science Foundation of China under grant U21A20520, grant 62325204, and grant 62072191, in part by the Key-Area Research and Development Program of Guangzhou City under grant 202206030009, and in part by the Guangdong Basic and Applied Basic Research Fund under grant 2023A1515030002 and grant 2024A1515011995.

\bibliographystyle{splncs04}

\end{document}